\documentclass[Afour,sageh,times]{sagej}

\usepackage{moreverb,url}

\usepackage[colorlinks,bookmarksopen,bookmarksnumbered,citecolor=blue,urlcolor=blue]{hyperref}

\usepackage{soul}

\newcommand\BibTeX{{\rmfamily B\kern-.05em \textsc{i\kern-.025em b}\kern-.08em
T\kern-.1667em\lower.7ex\hbox{E}\kern-.125emX}}

\def\journalname{The International Journal of Robotics Research}
\def\volumeyear{2024}

\usepackage{subcaption}
\usepackage{amsmath}
\usepackage{amssymb}
\usepackage{pifont}
\usepackage[table,dvipsnames]{xcolor}
\usepackage{arydshln}
\usepackage{placeins} %
\usepackage{threeparttable}
\usepackage{dirtree}
\usepackage{balance}
\usepackage{multirow}
\usepackage{booktabs}
\usepackage{tablefootnote}

\newcommand{\coolname}{\textit{WildScenes}} %

\newcommand{\eg}{\emph{e.g.},}
\newcommand{\ie}{\emph{i.e.},}
\newcommand{\etal}{\emph{et~al.}}
\newcommand{\cmark}{\ding{51}}%
\newcommand{\xmark}{\ding{55}}%

\newcommand{\symbolHt}{1.4em}
\newcommand{\forest}{%
  \begingroup\normalfont
  \includegraphics[height=\symbolHt]{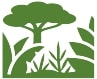}
  \endgroup
}
\newcommand{\city}{%
  \begingroup\normalfont
  \includegraphics[height=\symbolHt]{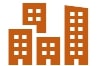}%
  \endgroup
}

\definecolor{mycolor}{RGB}{180, 211, 178}
\definecolor{mycolor_old}{rgb}{0.96,0.87,0.83}

\definecolor{bush}{RGB}{230, 25, 75}
\definecolor{dirt}{RGB}{60, 180, 75}
\definecolor{fence}{RGB}{0, 128, 128}
\definecolor{grass}{RGB}{128, 128, 128}
\definecolor{gravel}{RGB}{145, 30, 180}
\definecolor{log}{RGB}{128, 128, 0}
\definecolor{mud}{RGB}{255, 225, 25}
\definecolor{object}{RGB}{250, 190, 190}
\definecolor{other-terrain}{RGB}{70, 240, 240}
\definecolor{rock}{RGB}{170, 255, 195}
\definecolor{sky}{RGB}{0, 0, 128}
\definecolor{structure}{RGB}{170, 110, 40}
\definecolor{tree-foliage}{RGB}{210, 245, 60}
\definecolor{tree-trunk}{RGB}{240, 50, 230}
\definecolor{water}{RGB}{0, 130, 200}

\begin{document}

\runninghead{K. Vidanapathirana \etal: WildScenes}

\title{WildScenes: A Benchmark for 2D and 3D Semantic Segmentation in Large-scale Natural Environments}

\author{Kavisha Vidanapathirana\affilnum{1,2,}$^{\dagger}$,  Joshua Knights\affilnum{1,2,}$^{\dagger}$, Stephen Hausler\affilnum{1,}$^{*, \dagger}$, Mark Cox\affilnum{1}, Milad Ramezani\affilnum{1}, Jason Jooste\affilnum{1}, Ethan Griffiths\affilnum{1,2}, Shaheer Mohamed\affilnum{1,2}, Sridha Sridharan\affilnum{2}, Clinton Fookes\affilnum{2} and Peyman Moghadam\affilnum{1,2}}

\affiliation{\affilnum{1}CSIRO Robotics, \emph{firstname.lastname}@csiro.au\\
\affilnum{2} Queensland University of Technology, \emph{firstname.lastname}@qut.edu.au\\
$^{\dagger}$ These authors contributed equally}

\corrauth{$^{*}$ Stephen Hausler, CSIRO Robotics,
1 Technology Ct, Pullenvale,
Brisbane,
Queensland, Australia}

\email{stephen.hausler@csiro.au}

\begin{abstract}{
Recent progress in semantic scene understanding has primarily been enabled by the availability of semantically annotated bi-modal (camera and LiDAR) datasets in urban environments. However, such annotated datasets are also needed for natural, unstructured environments to enable semantic perception for applications, including conservation, search and rescue, environment monitoring, and agricultural automation. Therefore, we introduce \coolname{}, a bi-modal benchmark dataset consisting of multiple large-scale, sequential traversals in natural environments, including semantic annotations in high-resolution 2D images and dense 3D LiDAR point clouds, and accurate 6-DoF pose information. The data is (1) trajectory-centric with accurate localization and globally aligned point clouds, (2) calibrated and synchronized to support bi-modal training and inference, and (3) containing different natural environments over 6 months to support research on domain adaptation. 
Our 3D semantic labels are obtained via an efficient, automated process that transfers the human-annotated 2D labels from multiple views into 3D point cloud sequences, thus circumventing the need for expensive and time-consuming human annotation in 3D. 
We introduce benchmarks on 2D and 3D semantic segmentation and evaluate a variety of recent deep-learning techniques to demonstrate the challenges in semantic segmentation in natural environments. We propose train-val-test splits for standard benchmarks as well as domain adaptation benchmarks and utilize an automated split generation technique to ensure the balance of class label distributions. 
The \coolname{} benchmark webpage is \href{https://csiro-robotics.github.io/WildScenes/}{https://csiro-robotics.github.io/WildScenes}, and the data is publicly available at \href{https://data.csiro.au/collection/csiro:61541}{WildScenes Benchmark Dataset}.}
\end{abstract}

\maketitle

\section{Introduction}
\label{sec:intro}

For autonomous agents to operate beyond the structured and controlled environments of urban streets and warehouses, they require the ability to understand the natural world. Perception in natural environments consists of additional complexities as these environments generally contain highly irregular and unstructured elements, making them less predictable than structured environments. For robots to account for these complexities, they must perceive the environment at a fine-grained level. Fine-grained semantic scene understanding has enabled many applications in urban environments, including mapping, localization, object retrieval, and dynamic situational awareness. Research progress on these tasks has primarily been enabled by the availability of semantically annotated bi-modal (camera and LiDAR) datasets.

While there are many semantic segmentation datasets, most of them focus on structured environments such as outdoor urban areas~(\cite{behley2021towards,cordts2016cityscapes,zhou2017scene}), or indoor environments (\cite{silberman2012indoor,ruiz2017robot}). Consequently, there is a need for more large-scale semantic segmentation datasets in unstructured, natural environments. These types of environments pose several challenges beyond those encountered in more structured environments. Firstly, in an urban environment, it is clear to define a building or a road, yet the separation between classes can be less defined in a natural environment. For example, the distinction between \emph{dirt} and \emph{mud}, or between \emph{grass} and a \emph{shrub}. Secondly, the density (or clutter) of natural environments causes boundary ambiguity to occur due to the co-occurring nature of natural semantic classes (\eg{} tree leaves and trunks). These challenges make semantic segmentation incredibly challenging in these environments and make it all the more important for future research to focus on developing robust perception systems that can operate in these environments (\cite{borges2022survey}).

\begin{figure*}[t]
    \centering
    \includegraphics[width=1.99\columnwidth, trim=0cm 0.0cm 0.0cm 0.0cm,clip]{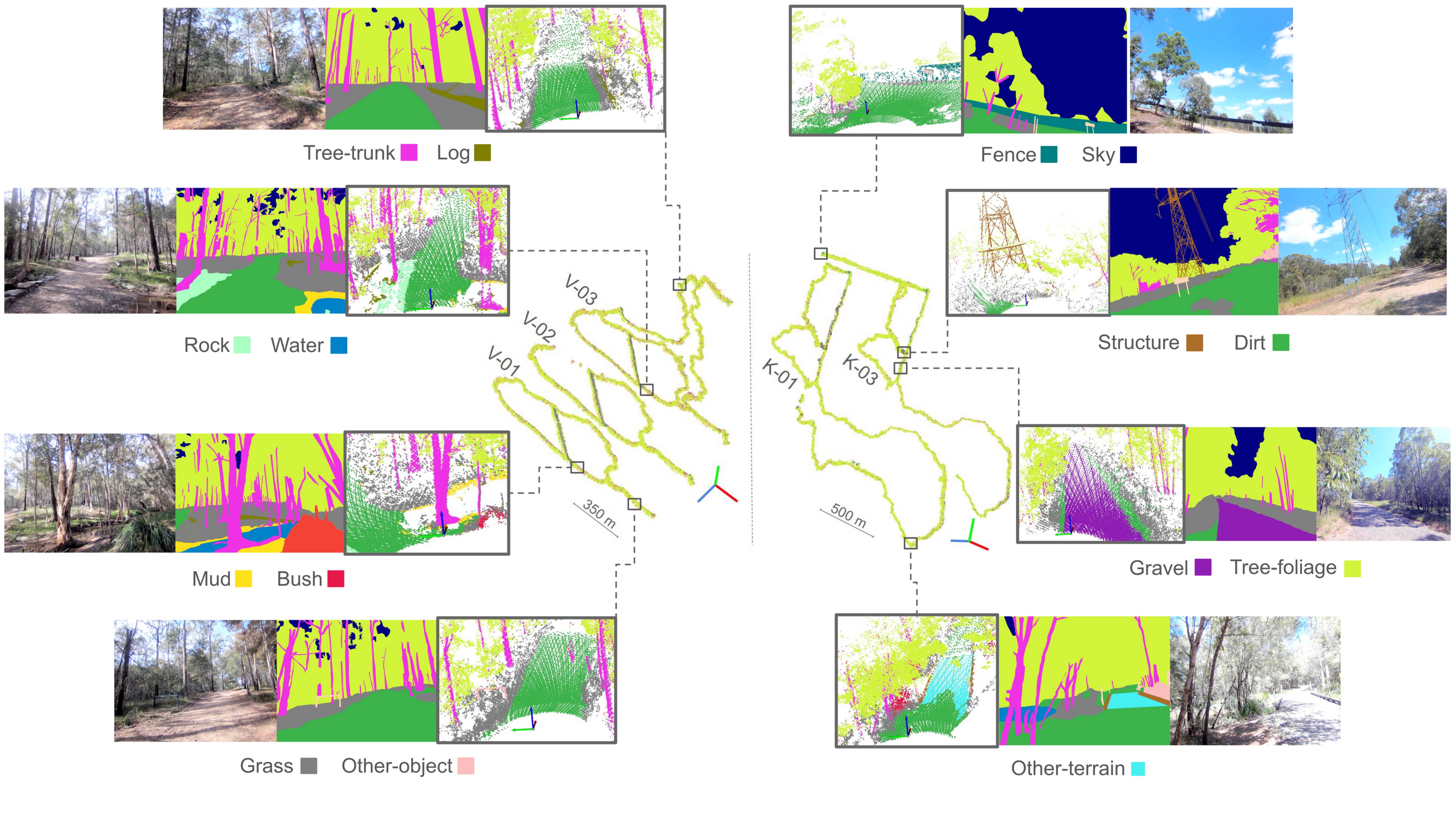}
    \caption{
    The \coolname{} benchmark consists of five large-scale traversals in two natural forest environments - Venman (V-01, V-02, V-03) and Karawatha (K-01, K-03). In the center of the figure, the traversals from each environment are depicted using the corresponding semantically annotated 3D global map of that traversal. The zoom-in views of example locations with prominent semantic classes are depicted. For each class example, three images depicting the 2D image, 2D semantic annotation, and 3D semantic point cloud of corresponding location and viewpoint are provided. 
    }
    \label{fig:hero}
    \vspace*{-0.2cm}
\end{figure*}

Therefore, we introduce \coolname{}, a bi-modal benchmark containing 2D and 3D semantic annotations in natural environments. \coolname{} comprises multiple sequential traverses through two distinct natural locations in Australia, including revisits after six months. 
This temporal and environmental diversity are essential properties for lifelong robotic applications to address the degradation in semantic inference that occurs due to either changes in environment or changes due to temporal and seasonal variations. \coolname{} provides long-term sequential synchronized 2D image and 3D point cloud data, and we provide accurate 2D semantic labels via human annotation (refer to Figure~\ref{fig:hero}). 

We also generate accurate 3D semantic point cloud labels using a LiDAR sensor along with significant post-processing. Our raw 3D data is generated using a spinning LiDAR sensor, which allows for a wide vertical field of view and is thus able to scan all objects in the environment, including tall trees. We then use 
a state-of-the-art LiDAR-inertial SLAM system
to generate an accurate 6-DoF trajectory and a globally consistent point cloud map for each traversal. Utilizing the globally consistent map and trajectory combined with precisely calibrated extrinsics and intrinsics, we are then able to accurately transfer the 2D semantic annotations from multiple views into the 3D point clouds in a manner that enforces the temporal and 2D-3D coherence of semantic labels. 

The main advantages of the \coolname{} benchmark include: scale (over 20km of sequential traversal over the course of six months), size (9,306 images and 12,148 point clouds), high 2D resolution ($2016 \times 1512$),  high 3D point density ($>$ 70,000 points per cloud) and accurate 6-DOF localization information. 
We provide: (1) rectified RGB images with 2D semantic annotations by human annotators. (2) 3D LiDAR point clouds, with motion distortion correction and 3D semantic annotations generated via efficient 2D label transfer. (3) Label distributions for 3D points to represent the semantic ambiguity in natural environments.

\begin{table*}[t]
    \caption{Comparison between recent and related semantic segmentation benchmark datasets, in urban and natural environments.
    }
    \vspace{-2mm}
    \begin{center}
    \tiny
    \resizebox{\linewidth}{!}{
    \begin{threeparttable}[t]
    \begin{tabular}{lcccccccc}
    \hline
    \multicolumn{1}{c}{\textbf{Dataset}}  &  \multicolumn{2}{c}{\textbf{Annotation}} &\multicolumn{2}{c}{\textbf{Diversity}} &\multicolumn{2}{c}{\textbf{Coherency}}  &\multicolumn{1}{c}{\textbf{Length}}   & \multicolumn{1}{c}{\textbf{\# Classes}} \\
      Urban: \city, Natural: \forest      & \textbf{Points}&\textbf{Pixels}& \textbf{Temporal}~\tnote{1} & \textbf{Environmental}~\tnote{2} & \textbf{Temporal}~\tnote{3} & \textbf{2D-3D}~\tnote{4} & (km)  & Natural (Total)~\tnote{5}\\ 
    \hline
    CityScapes \city & - & \cmark & - & \cmark & - & - & -  & 4 (30)\\
     SemanticKITTI \city  & \cmark&  - &  - & \cmark &  \cmark & - & 44  & 3 (19)\\
    KITTI-360  \city    & \cmark& \cmark& \cmark & \cmark &  \cmark & \cmark &  67  & 3 (19)\\
    Navya3DSeg \city & \cmark & - & - & \cmark & - & - & - & 2 (20)\\
    Panoptic nuScenes \city & \cmark & - & - & \cmark & - & - & - & 1 (16)\\ 
    SemanticSTF  \city   & \cmark&  - & - & - &  - & - &  -   & 3 (21)\\
    
    \hline 
    RUGD \forest     &  - & \cmark& - & \cmark & - &  - & -  & 9 (24)\\
    RELLIS-3D  \forest   & \cmark& \cmark& \cmark & - &  \cmark & - & 1.5 & 9 (20)\\
    \rowcolor{mycolor} \textbf{\coolname{} (Ours)} \forest    & \cmark& \cmark&\cmark& \cmark&\cmark& \cmark & 21 & 11 (15)\\
    \hline %
    \end{tabular}
    \begin{tablenotes}
    \item[1] `Temporal Diversity' denotes a revisit to the same environment at least six months apart to capture temporal and seasonal changes. 
    \item[2] `Environmental Diversity' denotes data collection in geographically separated (non-overlapping) environments with different characteristics. 
    \item[3] `Temporal Coherency' denotes semantic annotations in 3D, such that a point in space observed at different times will always have the same semantic label.
    \item[4] `2D-3D Coherency' denotes that semantic labels will remain consistent when projected from 3D to 2D using the provided calibration parameters. 
    \item[5] `\# Classes' includes only the subset of classes that have adequate sample size for evaluation/benchmarking purposes.
    \item[6] The `dash' symbol (-) denotes the absence of a particular property (noted in column title) or the lack of adequate information to determine its presence/value. 
    \end{tablenotes}
    \end{threeparttable}
    }
    \label{tab:comparison_table}
    \end{center}
    \vspace{-0.5cm}
    \end{table*}

Additionally, our benchmark dataset involves traversing through very dense and rough terrain, which could aid research on autonomous systems that need to operate in remote, difficult-to-access locations. To enable research on semantic domain adaptation in natural environments, we provide (1) traversals in geographically separated environments to capture a different sample distribution for each of our semantic classes and (2) repeat traversals in the same environment with a 6-month time gap to capture distribution shift due to temporal and seasonal changes.
In summary:

\begin{itemize}
    \item We introduce the \coolname{} benchmark, which contains synchronized 2D and 3D sequential, dense semantic annotations of multiple large-scale traversals in unstructured natural environments. 
    \item We provide a benchmark for semantic scene understanding in natural environments for 2D and 3D tasks. The benchmark contains train-val-test splits optimized to balance the class label distribution, including separate splits for domain adaptation downstream tasks.
    \item We provide a strategy for generating dense 3D labels without human annotators (named \emph{LabelCloud}), utilizing geometric projection from 2D labeled images with a robust visibility check. Additionally, this method generates a histogram of label assignments per point, which could be used in uncertainty-aware semantic segmentation algorithms.
\end{itemize}

\section{Related Work}
\label{sec:rel_work}

Semantic segmentation in 2D and 3D aims to assign pixel/point-wise labels for images and point clouds for scene understanding. There are many established datasets and learning-based methods developed for urban scene understanding, including SemanticKITTI (\cite{behley2021towards}), Cityscapes (\cite{cordts2016cityscapes}), KITTI-360 (\cite{liao2022kitti}), Panoptic nuScenes (\cite{fong2022panoptic}), Boreas (\cite{burnett2023boreas}), SemanticPOSS (\cite{pan2020semanticposs}), Rope3D (\cite{ye2022rope3d}), Navya3DSeg (\cite{almin2023navya3dseg}) and SemanticSTF (\cite{xiao20233d}), with an overview of the most prominent semantic segmentation datasets provided in Table \ref{tab:comparison_table}.  These datasets focus mainly on applications and scenarios for autonomous driving in urban environments.  A semantic understanding of the environment is key to path planning and navigation in natural environments, where traversability is dependent on a wealth of information beyond the 3D geometry of the scene.  For example, a legged platform that could navigate a flat stretch of grass or dirt with ease could experience a catastrophic failure attempting to navigate a stretch of mud or sand. Furthermore, natural environments can also undergo significant changes over short periods of time, due to seasonal changes in vegetation, or the effect of climate change. However, compared to urban environments, there is only a limited number of works focusing on natural/unstructured environments (\cite{dokania2023idd,wigness2019rugd, min2022orfd,jiang2021rellis,valada2017deep,marzoa2023magro,knights2023wildplaces}).

Several early datasets for perception in natural environments initiated a discussion around this area with datasets that contain very limited semantic annotations
as seen in (\cite{maturana2018real,metzger2021fine}).  TartanDrive (\cite{triest2022tartandrive}) lacks semantic annotations for 2D and 3D segmentation tasks, while providing off-road driving interactions with seven sensing modalities.  The \textit{Wild-Places} benchmark (\cite{knights2023wildplaces}) provides accurate 6-DoF submap poses for eight LiDAR sequences across two large-scale natural environments for benchmarking the task of LiDAR place recognition, but does not provide any semantic labels for training or evaluating semantic segmentation.  RUGD~(\cite{wigness2019rugd}) offers a large-scale dataset collected in off-road terrains but only provides RGB images.  Ideally, a large-scale dataset for semantic segmentation should contain annotations for both 2D images and 3D point clouds to allow exploration of the advantages of both modalities - for example, the rich color and textural information from 2D images and the 3D geometric information from LiDAR point clouds - can be used to enhance the semantic understanding of an environment.  ORFD~(\cite{min2022orfd}) introduces a bi-modal dataset for free-space and traversability detection in off-road scenarios under various weather conditions, but its semantic labels consist of only three classes: free space, traversable, and non-traversable.  The closest existing dataset to this work is RELLIS-3D~ (\cite{jiang2021rellis}), a bi-modal dataset that provides both RGB and LiDAR annotations for semantic segmentation in natural environments.  However, RELLIS-3D only covers a distance of 1.5km, and due to being collected by a robotic platform, the traversals present in the dataset are limited to fairly wide open trails and do not capture dense forest trails. 

To address these limitations, we introduce \coolname{}, a large-scale bi-modal benchmark with dense and sequential RGB and LiDAR annotations. Table \ref{tab:comparison_table} contains an overview and comparison of \coolname{} against other prominent 2D and 3D semantic segmentation benchmark datasets in both urban and natural environments.  \coolname{} contains 9,306 annotated images and 12,148 annotated submaps, spread over five traversals of the two natural environments explored in the \textit{Wild-Places} benchmark (\cite{knights2023wildplaces}) as outlined in Table \ref{tab:ws_seq_details}.  The dataset was collected by a human operator carrying a handheld multimodal sensor payload, enabling the capture of areas in dense forest trails inaccessible to vehicles or robotic platforms. To the best of our knowledge, such dense forest environments with sections of terrain untraversable to the current mobile robots are not represented in existing semantic segmentation datasets.

In addition, \coolname{} represents temporal domain shifts of more than six months in two different environments, ensuring a high degree of temporal and ecological diversity in the data. We provide benchmark splits for testing the domain adaptation capabilities of networks that account for this temporal and environmental diversity.
While domain adaptation has received much attention in the 2D domain, it has only recently become popular for 3D semantic segmentation (\cite{saltori2023walking, sanchez2023domain, jiang2021lidarnet, sanchez2023parisluco3d, knights2023geoadapt}). While current works primarily explore domain adaptation between different sensors and places, our dataset has a unique representation of domain shifts in the same environment due to the change in the characteristics of natural classes across time. We hope that our data and benchmarks will provide a platform for addressing the challenge of domain adaptation - which is a crucial ability for lifelong autonomy.

Finally, we provide accurate 6-DoF ground truth poses (with precise calibration and synchronization) for our annotated 2D images and 3D clouds to allow \coolname{} to be used to investigate how temporal coherency (\cite{sun2022coarse, nunes2023temporal, wu2023spatiotemporal, baghbaderani2024temporally}) and multi-modal fusion (\cite{krispel2020fuseseg, zhuang2021perception, yan20222dpass}) can enhance semantic segmentation performance.

\section{\coolname{} Benchmark Dataset}
\label{sec:problem}

\begin{table}[t]
    \centering
    \caption{The five traversals of the \coolname{}.}
    \resizebox{0.95\linewidth}{!}{
    \begin{tabular}{ccccccc}
        \hline 
        \textbf{Environment} & & \textbf{Date} & \textbf{Length} & \textbf{Duration} & \textbf{Images} & \textbf{Submaps}\\
        \hline 
        \multirow{3}{*}{\hfil Venman}
        & V-01 & June 2021  & 2.64 km & 39m & 743 & 1080 \\
        & V-02 & June 2021 & 2.64 km & 38m & 833 & 1100 \\
        & V-03 & Dec 2021 & 4.59 km & 1h 11m & 1845 & 2407\\
        \hline 
        \multirow{2}{*}{\hfil Karawatha}
        & K-01 & June 2021 & 5.14 km & 1h 14m & 1972 & 2271 \\
        & K-03 & Dec 2021 & 6.27 km & 2h 7m & 3913 & 5290 \\
        \hline 
        Total & 5 & 6 months & 21.28 km & 5h 49m & 9306 & 12148 \\
        \hline 
    \end{tabular}
    }
    \label{tab:ws_seq_details}
\end{table}

\begin{figure}[]
    \centering
    \includegraphics[width=0.99\columnwidth]{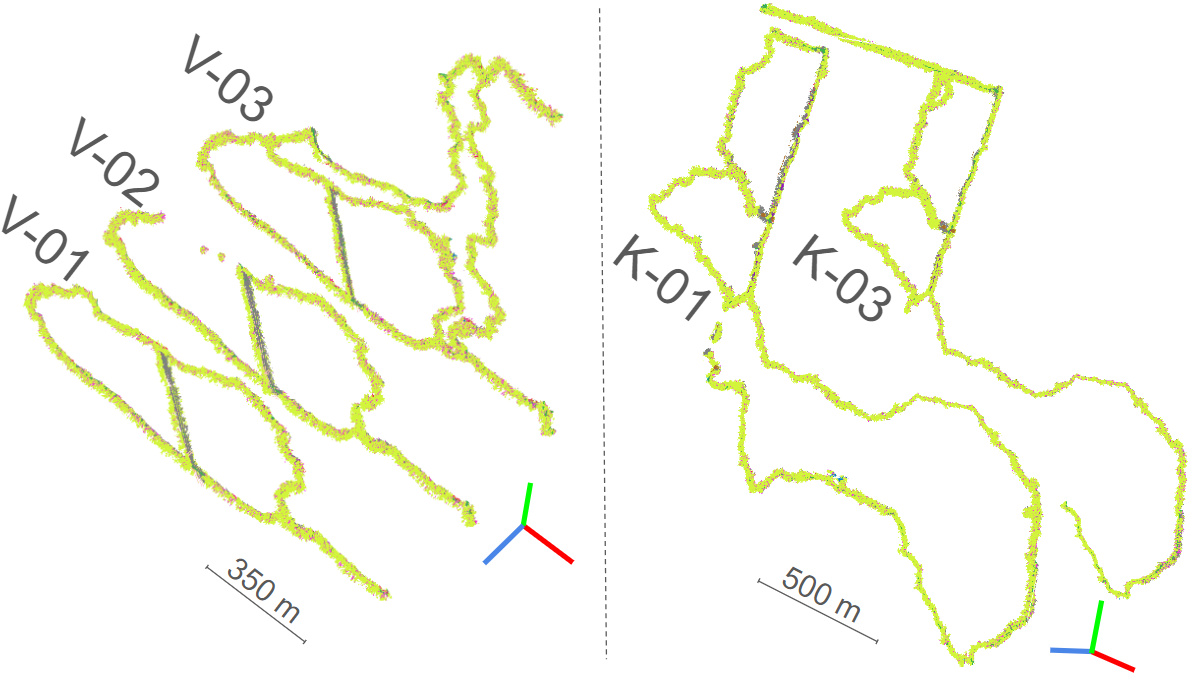}%
    \caption{The 3D semantic maps of the five traversals. The \coolname{} contains repeat traversals of two natural environments, Venman (V-01, V-02, V-03) (left) and Karawatha (K-01, K-03) (right). 
    }
    \label{fig:trajs}
\end{figure}

\begin{figure}[t!]
    \centering
    \includegraphics[width=0.99\columnwidth]{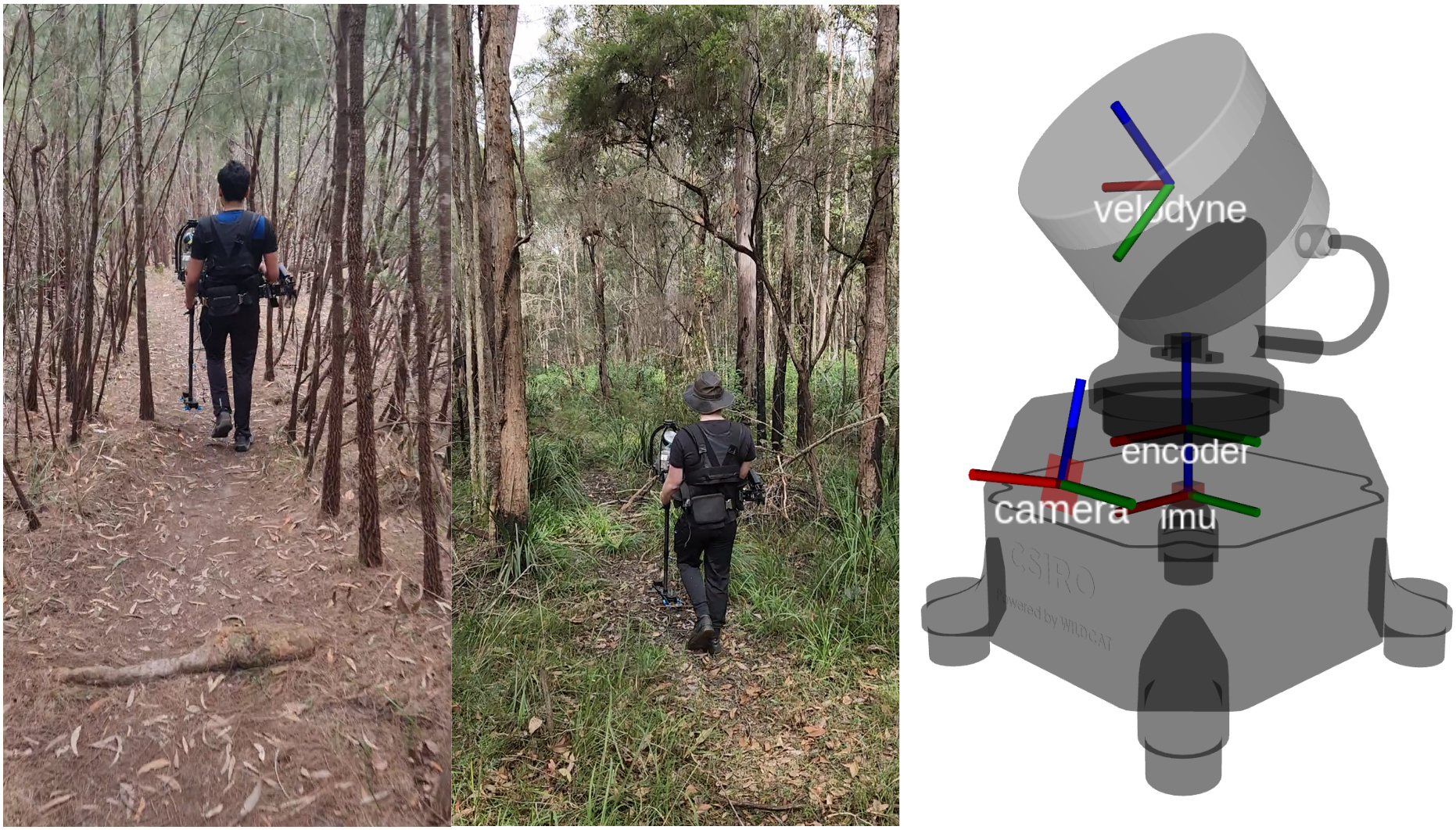}
    \caption{Data collection campaign depicting the dense forest trails of Karawatha and Venman, respectively (left). The sensor payload comprises a spinning LiDAR sensor, encoder, IMU, GPS, and camera (right).
    }
    \label{fig:datacollection}
\end{figure}

The dataset used for \coolname{} benchmark is a multi-modal collection of traversals within Australian forests, allowing for a range of computer vision and robotic applications in natural environments. The \coolname{} is divided into five sequences across two forest locations: Venman National Park and Karawatha Forest Park, Brisbane, Australia. These sequences are across different physical locations and also across different times. Please see Table~\ref{tab:ws_seq_details} and Figure~\ref{fig:trajs} for more details on the dataset traversals. The data was collected by walking through these locations with a portable, handheld sensor payload, as shown in Figure~\ref{fig:datacollection}. For each traverse, we provide an accurate 6-DoF ground truth pose, manually annotated 2D semantic segmentation images, and generated 3D semantic segmentation point clouds. In total, \coolname{} provides 9,306 images of $2016\times1512$ resolution and 12,148 associated point cloud submaps with greater than 70,000 annotated points per submap (on average). The number of points per submap can vary due to the spatial density of the environment, \ie{} due to the proportion of trees versus sky.

\begin{table}[t!]
\caption{Sensor specifications used in \coolname{}.}
\vspace{-0.4cm}
\tiny
\begin{center}
\resizebox{\linewidth}{!}{
\begin{tabular}{llcl}
\hline
\textbf{Sensor} & \textbf{Model} & \textbf{Rate (Hz)} & \textbf{Specifications} \\
\hline
LiDAR$^*$ & VLP-16 & 20 & 16 Channels \\
& & & 120m Range \\
Camera x4 & e-CAM130A CUXVR & 15 & $94.9^{\circ}$H FOV \\
& & & $71.2^{\circ}$V FOV \\
IMU & 3DM-Cv5-25 & 100 & 9-DoF \\
GPS & Ublox-Neo-M8N& 1& \\
\hline
\end{tabular}}
\label{tab:sensor_specs}
    \scriptsize
    \begin{tablenotes}
    \item[*] *Our mechanical design allows $120$ degrees vertical FoV (see  Section~\ref{sec:problem}).
    \end{tablenotes}
\end{center}    
\vspace{-0.25cm}
\end{table}

Our portable sensor setup includes a Velodyne Puck LiDAR sensor (consisting of 16 beams) attached to a brushless DC motor, rotating at a frequency of 0.5 Hz around the z-axis. This strategy is utilized to increase the vertical field-of-view and the 3D point density - given the Puck's restricted vertical field of view (30 degrees), it is inclined at an angle of 45 degrees on the DC motor and rotated around an external axis. This setup enables LiDAR scans with a 120 degrees vertical Field of View (FoV), making it suitable for comprehensive top-to-bottom mapping of features such as trees. Additionally, it features a Microstrain 3DM-CV5-25 9-DoF IMU, a Ublox GPS antenna, and a Nvidia Jetson AGX Xavier. Pulse Per Second (PPS) is utilized to achieve sub-microsecond time synchronization accuracy among the sensors. The sensor payload is further equipped with four cameras for visual perception; however, we only use the front camera for annotation. We provide a complete summary of all sensor specifications in Table~\ref{tab:sensor_specs}. %

To provide an accurate localization and mapping ground truth, we employ the LiDAR-inertial SLAM system Wildcat (\cite{ramezani2022wildcat}) in which 6-DoF poses 
are optimized within a sliding window of LiDAR and inertial measurements captured in time. Our odometry system is devised to merge asynchronous IMU readings and LiDAR scans effectively through continuous-time trajectory representations (\cite{bosse2009continuous,furgale2012continuous,droeschel2018icra, park2021elasticity}). A primary benefit of continuous-time trajectory representation is to query corrected positions of LiDAR points at their timestamps, alleviating map distortion caused by the sensor's motion. This is critical due to the extreme motions of the handheld mobile sensors. %

\begin{figure}[t]
    \centering
    \includegraphics[width=0.99\columnwidth, trim=0cm 0cm 0.0cm 0.0cm,clip]{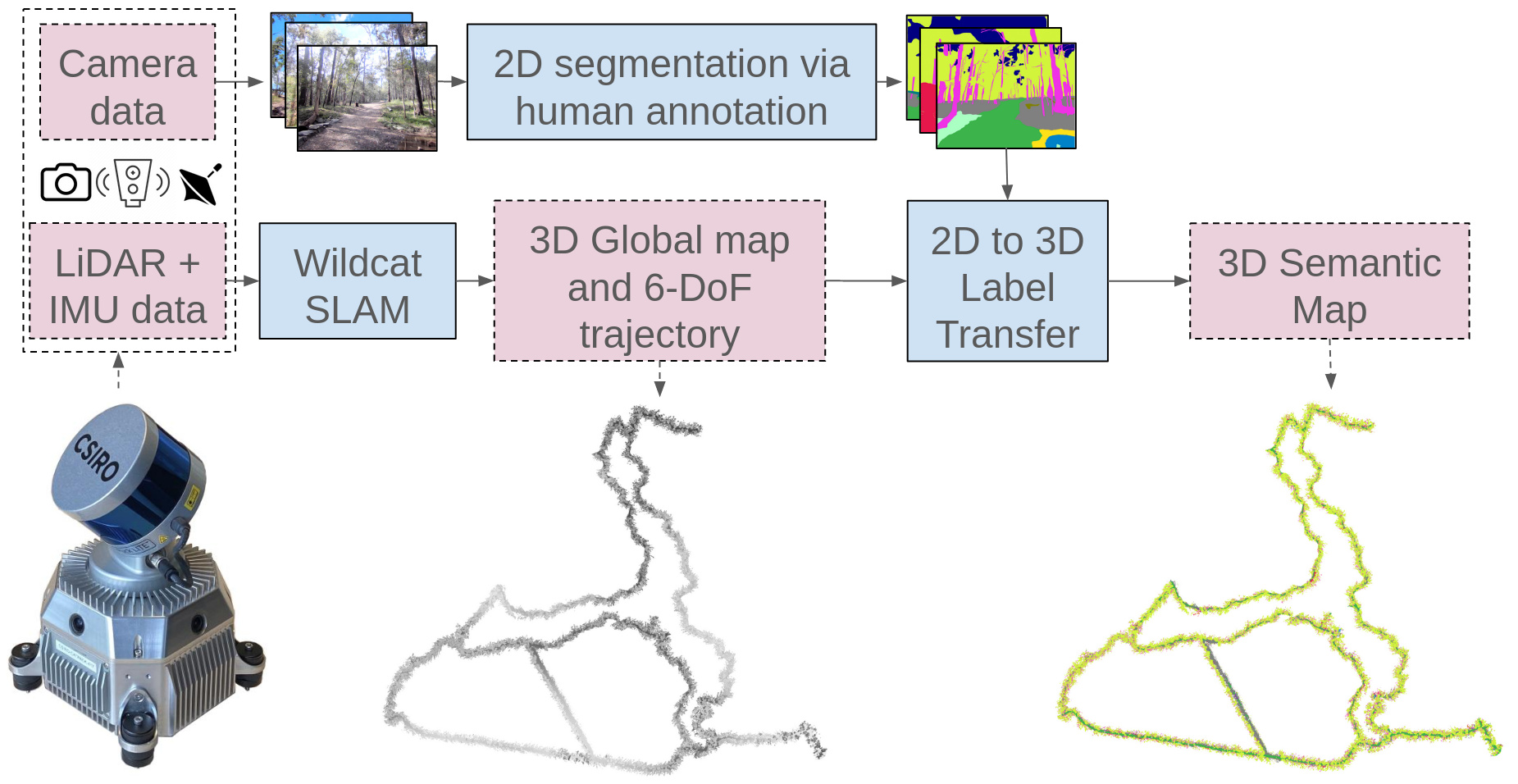} %
    \caption{Overview of the \textit{LabelCloud} pipeline for generaing a 3D semantic map. We use Wildcat SLAM to calculate the trajectory and global map. Then, after annotating the 2D images, we perform label transfer from 2D images across multiple frames into 3D, utilizing the 6-DOF trajectory, to produce our 3D semantic point cloud. }
    \label{fig:pipeline}
\end{figure}

To remove drift over time and generate a globally consistent map, we further incorporate GPS measurements into an offline bundle adjustment to optimize localization and mapping across the entire collection of IMU and LiDAR measurements, along with loop-closure constraints derived from a mechanism of loop-closure detection in revisit places. The bundle adjustment and the employed continuous-time trajectory representation allow the provision of a near-ground-truth trajectory and an undistorted map of the environment. This process allows us to release 3D point clouds that are globally aligned and consistent across an entire traverse, not just frame-to-frame. When creating the 3D point clouds, we apply a self-strike mask, a filter designed to exclude
points that hit the person carrying the device. The radius set for the self-strike mask during our data
post-processing is 2m.

For the purpose of annotation, we sample a new image frame from the video stream for every five meters traveled or after every cumulative five degrees of rotation in the heading angle of the payload using the 6-DOF estimated global trajectory from SLAM. Since the sensor motion of the handheld sensor and the walking patterns of individuals can vary a lot, we employed this trajectory-centric sampling regime, as opposed to the commonly used equal temporal interval-based sampling, to ensure consistent sampling of all regions covered in the trajectory. 

\begin{figure}[t]
    \centering
    \begin{subfigure}[b]{\columnwidth}
        \includegraphics[width=0.99\columnwidth, trim=0cm 0cm 0.2cm 0.2cm,clip]{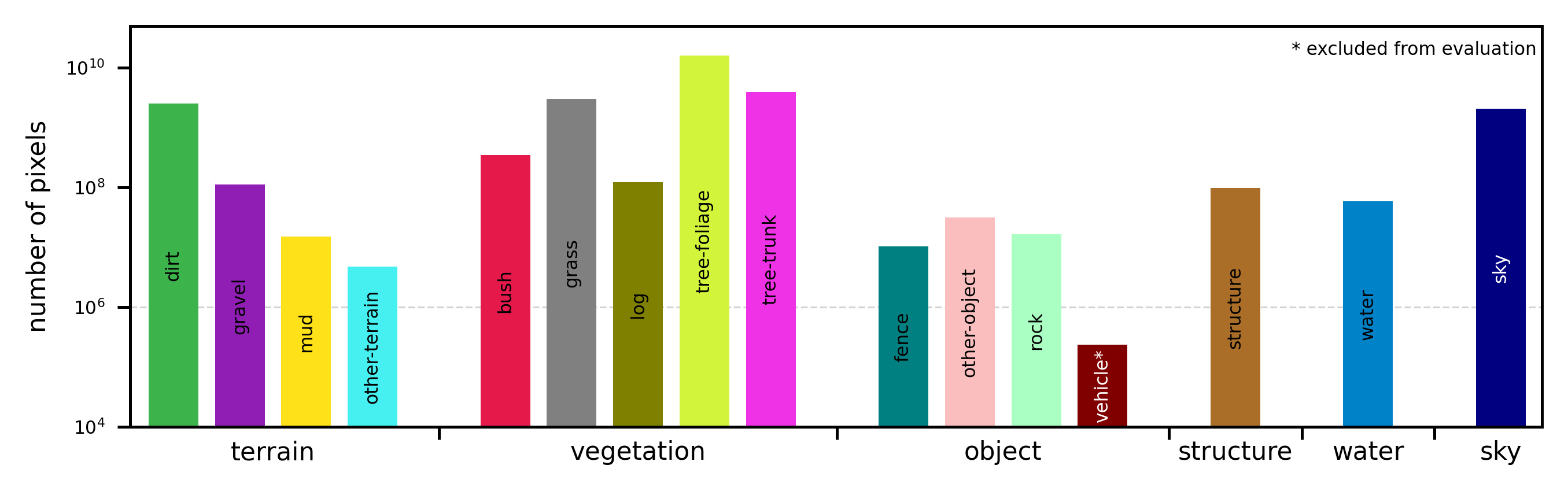}
    \end{subfigure}
        \begin{subfigure}[b]{\columnwidth}
        \includegraphics[width=0.99\columnwidth, trim=0cm 0cm 0.2cm 0.2cm,clip]{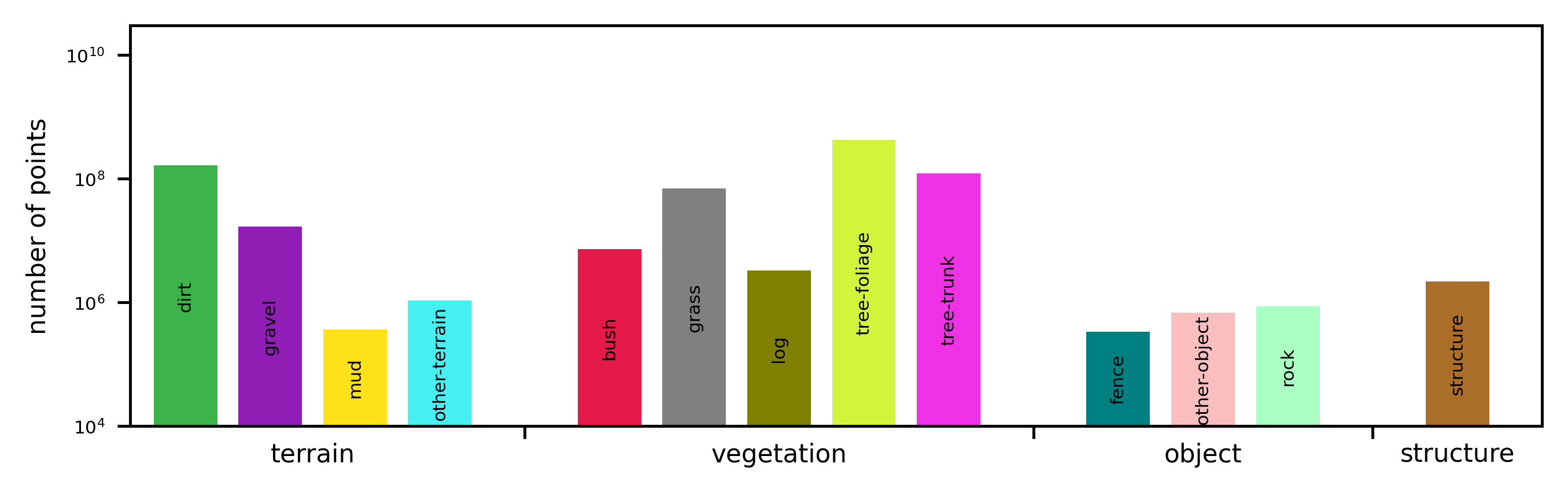}
    \end{subfigure}
    \caption{2D (top) and 3D (bottom) label counts of \coolname{}. The dashed line in the 2D counts represents the threshold for exclusion of a class for evaluation.
    }
    \label{fig:classlabeldist}
\end{figure}

For each sample image, we generate a corresponding LiDAR submap (from the global map noted earlier) by accumulating points within a 45-meter radius from the sensor frame and one second before and after the image timestamp. With precise sensor calibration and the retrieval of sensor 6-DOF pose from the SLAM-estimated trajectory, the projection of LiDAR submaps onto corresponding sample images is achievable. Additionally, we use camera intrinsic parameters to rectify sample images for the labeling process.

In summary, \coolname{} is created using a pipeline of steps, from localization using LiDAR-inertial SLAM to human annotation of sampled images. After producing a trajectory-centric and globally aligned point cloud, we perform multi-frame label transfer from 2D into 3D to produce an accurate 3D semantic annotated map. Figure~\ref{fig:pipeline} summarises this process.

\subsection{2D Semantic Annotations}

We provide manually annotated semantic segmentations for every sampled image in \coolname{}, dividing the observed scene into a collection of different natural-scene classes. \coolname{} comprises 15 different classes for the benchmark. Our class list is designed for natural environments and contains precise separation of vegetation types including, for example, \emph{tree-foliage} (leaves) versus \emph{tree-trunk}, and the distinction between different terrain features such as \emph{dirt} and \emph{mud} (as shown in Figure~\ref{fig:classlabeldist}). Further details about our class list are provided in the \coolname{} Supplementary Material section of this paper (Table. \ref{tab:semdesc}).

Several challenges arise when attempting to annotate unstructured, natural environments with such class specificity. In the wild, it can be hard to differentiate similar terrains or objects, such as dirt, mud, and gravel, depending on lighting conditions. Additionally, boundary ambiguity is a major issue due to the overlap of branches, leaves, bushes, etc. To mitigate these challenges, we follow a coarse-to-fine annotation approach to efficiently complete large-scale annotation while ensuring label quality and consistency with further refinements. 

The first round of annotation produced coarse semantic labels by randomly distributing images between a group of experienced annotators. These annotations underwent multiple rounds of auditing to correct major errors and missing labels. All annotators used the same ontology for annotation, but due to the aforementioned ambiguity of classes and boundaries in natural environments, there were some inconsistencies in labels between different annotators.  %

To ensure consistent annotation, a final round of fine-grained auditing was done by a single trained annotator for approximately 250 hours of annotation time. This audit focused on ensuring temporal consistency within sequences and enforcing class uniformity of features, as it was common for an ambiguous object to have differing class labels both intra-sequence and inter-sequence. This process significantly improved the quality and consistency of the coarse semantic annotations.

\subsection{3D Annotations using \textit{LabelCloud}} 
\label{sec:3dannot}

Compared to the manual annotation of 2D images, dense, cluttered, and unstructured natural environments make point-wise annotation of 3D point clouds very challenging. Forest environments have challenges, including occlusions and overlay between different semantic elements in the point cloud, the high spatial frequency of natural elements, and the inherent boundary ambiguities between different natural features such as dirt and mud. Therefore, the process of 3D annotating natural environments is more extensive in terms of both time and cost, resulting in practical infeasibility for large-scale datasets in dense forest environments.

To better facilitate large-scale 3D semantic annotation of point clouds, we propose a technique (named \emph{LabelCloud}) for accurate and robust transfer of 2D semantic labels from multiple viewpoints onto a 3D point cloud. This process is depicted in Figure \ref{fig:pipeline}. \emph{LabelCloud} is inspired by the 3D point colorization algorithm (\cite{vechersky2018paintcloud}), where each 3D point is assigned an RGB color according to their projections from image frames to point clouds. In this work, \emph{LabelCloud} estimates the full distribution of label counts per 3D point. It also provides the mode over the distribution of 2D observations to find the most commonly observed labels.

A critical component of transferring data from a pixel in an image $\mathbf{I}_C(\tau)$ acquired by camera $C$ at time $\tau$ to a 3D point in a point cloud is ensuring the 3D point is visible in the image plane. To perform this task, we employ a multi-step process. The first step identifies the 3D points that are within a specific range of the camera $C$. This ensures that far-away observations of 3D structures where the image resolution is poor do not contribute to the final assessment of the 3D point. The next filter selects 3D points that project to valid pixel locations inside the image $\mathbf{I}_C(\tau)$. The surface normal of a 3D point is then used to ensure that the 3D point is observable by the camera. The surface normal of a 3D point is calculated from the covariance matrix constructed from the neighboring 3D points. The eigenvector of the covariance matrix with the smallest eigenvalue is assumed to be the surface normal. The ambiguity in the direction of the surface normal is resolved by considering the position of the sensor payload at the time the 3D point was first observed.

\begin{figure}[t]
    \centering
    \includegraphics[width=0.90\columnwidth]{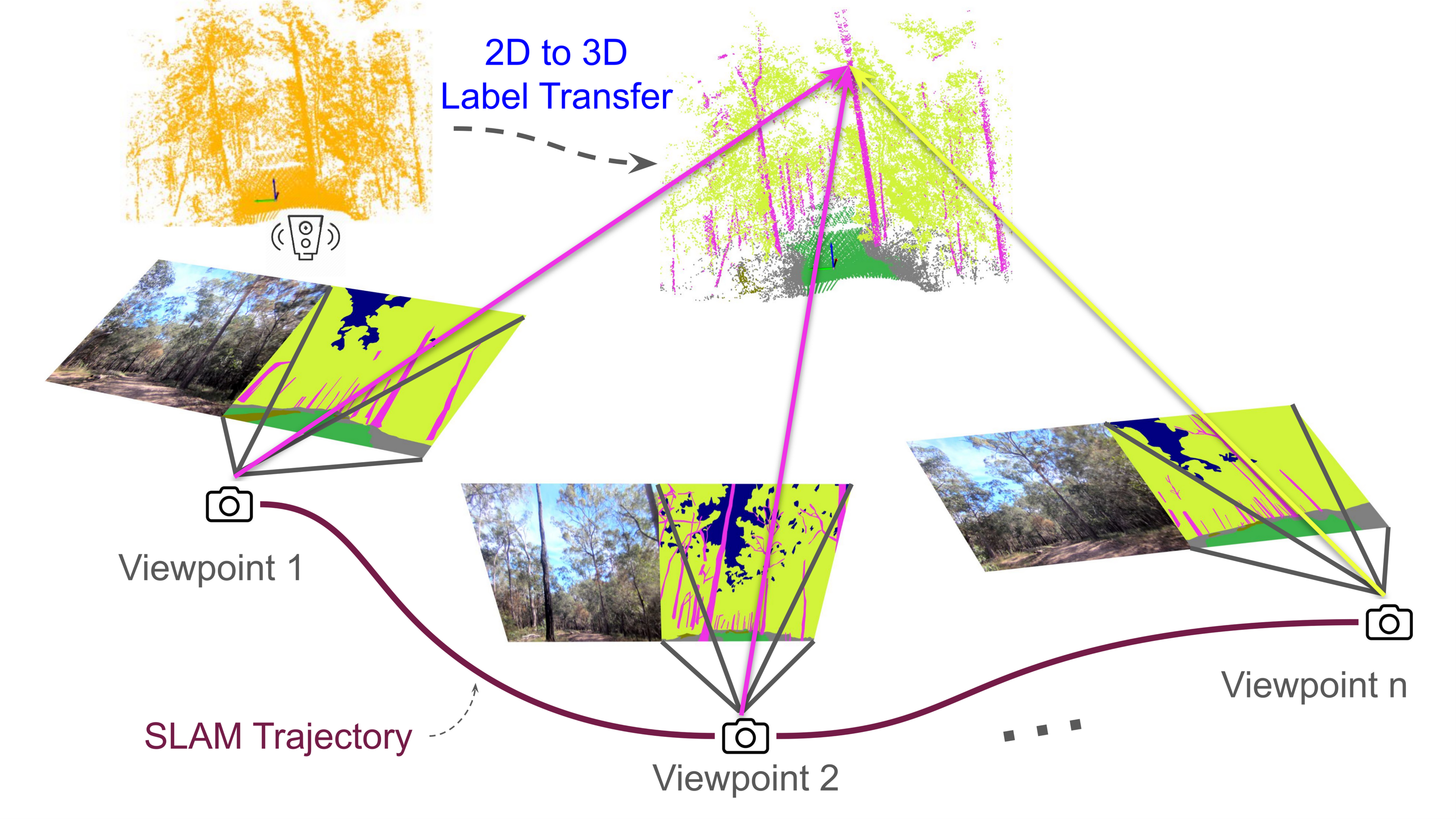}%
    \caption{
    An illustration of how 3D semantic labels are generated from the 2D labelled images. We first compute the set of images which observe a single 3D point (shown as viewpoint 1-n). A histogram of labels for the 3D point is then calculated by projecting the 3D point on to each image and recording the associated 2D label. The 2D label with the highest number of observations is then transferred to the 3D point. This example shows the label transfer for a point on a tree trunk. We see two votes for the label `\textcolor{tree-trunk}{tree-trunk}' and one vote for the label `\textcolor{tree-foliage}{tree-foliage}'.}
    \label{fig:paintcloud}
\end{figure}

The final step of identifying the visible 3D points addresses the problem of determining 3D structures that occlude other 3D structures. This is a challenging problem as 3D points have no volume, and it is improbable for two or more 3D points to lie on the same ray passing through the camera center. To address the problem, we employ the use of the generalized hidden point removal operator (\cite{katz2015ghpr}). The operator identifies visible points by performing a spherical reflection such that the order of points by distance is reversed \ie{} 3D points that are closest to the camera center become the furthest and vice versa. The 3D points on the convex hull of the reflected point cloud are classified as visible. The function used to perform the spherical reflection governs how visibility is determined, and the inverse reflection of the edges connecting the 3D points on the convex hull represents the hallucinated 3D surface of the visible points. In this work, we use the exponential inversion kernel to perform the reflection due to its scale-invariant properties (\cite{vechersky2018paintcloud}).

Having identified the 3D points that are visible in the image $\mathbf{I}_C(\tau)$, it is now straightforward to augment information about each visible 3D point with its corresponding image pixel. We begin by projecting the set of visible 3D points onto the image, then returning the observed label at the corresponding pixel location. We then repeat this process for all images, and if a given 3D point is visible in multiple images, then we aggregate all 2D labels that we acquire from projection. To find the final label per point, we find the most commonly observed label, \ie{} the mode over the distribution of 2D observations. A visualization of this process is provided in Figure \ref{fig:paintcloud}. Note that if a 3D point is not visible in any image, we drop this point from our labelled point cloud output.

Furthermore, we also record the full distribution of 2D observations for each 3D point, providing a histogram of label observations per point. This provides several benefits: firstly, it provides an opportunity to measure the consistency of the 2D semantic labels for a specific 3D point. Secondly, the most probable semantic label for each 3D point can still be used to train 3D semantic segmentation algorithms.  Finally, by providing the full distribution of label observations, we hope to provide an avenue for future researchers to use this data to explore novel research areas such as uncertainty-aware (\cite{sirohi2023uncertainty,cortinhal2020salsanext}) or multi-label semantic segmentation (\cite{zhu2019improving}).  Hereafter, we refer to this multi-view aggregation of observations as our \emph{label histogram}.

Therefore, for each 3D point, we generate a label histogram $P^i$ for the $i^{th}$ point, where $\left|P^i\right|=C$ and $C$ is the number of classes in \coolname{}. Such a distribution could have a myriad of future applications, such as label distribution learning or uncertainty-aware perception. We provide a quantitative analysis of this label ambiguity in our results and discussion section. 

\subsection{Split Generation}
\label{sec:split}

As \coolname{} was recorded from sequential traversals of natural environments, there exists a risk of geographical proximity between the train/test/validation sets. Additionally, it is important to have train, validation, and test sets with a uniform class distribution.

To address this, we developed a split generation procedure that adds buffer regions between sets while also ensuring a good class distribution. As our 3D submaps have a radius of $45\text{m}$, we added buffer regions such that there is a minimum distance of $45\text{m}$ between samples from different sets. The buffer regions are designed to ensure no visual overlap between train/val/test splits (2D and 3D). %

We used a modified version of the split generation procedure proposed by \cite{navya} to generate an optimized split that satisfied our requirements.
In summary, the algorithm generates a large number of candidate splits by randomly assigning chunks of the trajectory into candidate sets and then selects the best split based on a number of metrics and constraints. 
As \coolname{} contains a number of sequences across different times, samples were grouped based on their 2D (x,y) coordinates with k-means clustering for $K=50$. This was performed to bias the generation of candidate splits towards less interleaving. The parameter $k$ represents a tradeoff where smaller $K$ allows a larger space of candidate splits but biases the candidates towards highly interleaved splits with large numbers of images lost to the buffer. 

Subsequently, $1000$ candidate splits were generated with random initialization, each satisfying the constraint that all (train/val/test) sets had at least one instance of each class. We used the Label Distribution ($m_{LD}$), the Inverse Frequency Weighted Label Distribution ($m_{IF}$), and the Label KL Divergence ($m_{KL}$) to calculate a fitness score for each of these candidates. In summary, these metrics estimate the divergence between the class distribution in a subset of a given split compared to the class distribution of the full dataset. Ideally, the distribution of class counts in a given split should match the class distribution of the full dataset.

Unlike the work in \cite{navya}, because of the need to include buffer regions, we added the Silhouette Coefficient ($m_{SC}$) (\cite{rousseeuw1987silhouettes}) as an additional metric. $m_{SC}$ calculates how tightly grouped each of the sets is in metric space with respect to the other sets. $m_{SC}$ is defined as:
\begin{equation}
    m_{SC} = \frac{1}{N} \sum_i \frac{b_i - a_i}{\max \left(a_i,b_i\right)},
\end{equation}

where $a_i$ is the mean intra-split distance and $b_i$ is the mean nearest-split distance for sample $i$, for all samples $N$ in each split. A high $m_{SC}$ value means the split contains samples from a similar metric location, while a low $m_{SC}$ value means there is a large amount of interleaving between sets.

The final split quality was calculated using a normalized and weighted combination of the aforementioned metrics:
\begin{multline}
    S^r = w_{LD}\frac{m^r_{LD} - \mu\left(m_{LD}\right)}{\sigma\left(m_{LD}\right)}
    + w_{IF}\frac{m^r_{IF} - \mu\left(m_{IF}\right)}{\sigma\left(m_{IF}\right)} \\
    + w_{KL}\frac{m^r_{KL} - \mu\left(m_{KL}\right)}{\sigma\left(m_{KL}\right)} - w_{SC}\frac{m^r_{SC} - \mu\left(m_{SC}\right)}{\sigma\left(m_{SC}\right)},
\end{multline}

where $r$ is the index of the random split in $R$ and $|R| = 1000$. 
The optimal split was found by finding the random split with the minimum score $S^r$, using the 2D labels as the input to the algorithm. We set $w_{LD} = 1$, $w_{IF} = 1$, $w_{KL} = 1$ and $w_{SC} = 2$. We prioritize the Silhouette Coefficient metric because we found that excessive generation of buffer regions significantly reduced the size of our train/val/test sets. %

\begin{figure}[t]
    \centering
    \includegraphics[width=0.99\columnwidth, trim=0.0cm 1.0cm 0.0cm 0.0cm,clip]{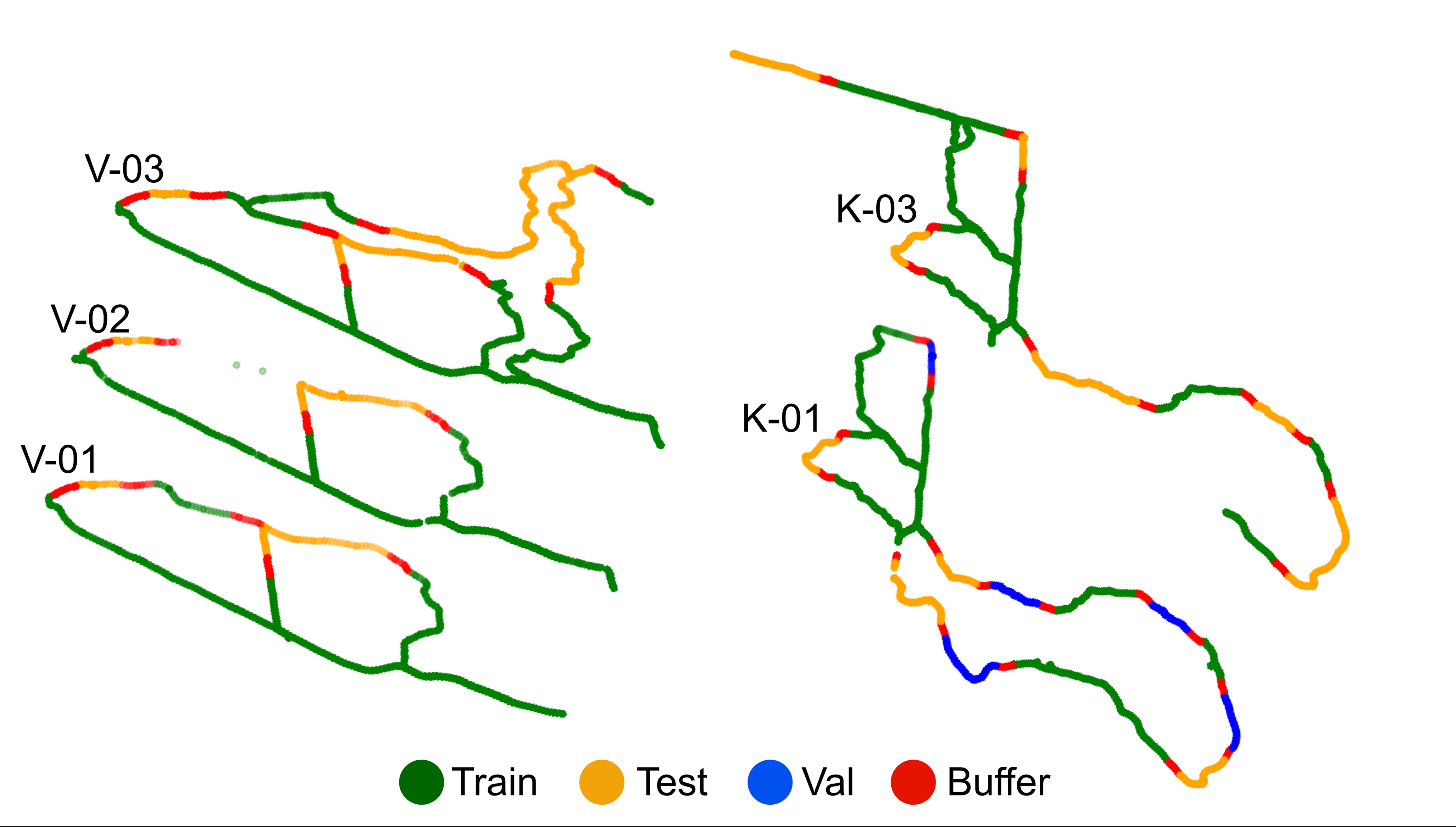}%
    \caption{Metric maps of the \coolname{} benchmark dataset showing the geographical distribution of train/val/test sets and buffer regions. 
    }
    \label{fig:splitplot}
\end{figure}

\begin{figure}[t]
    \centering
    \begin{subfigure}{\columnwidth}
        \includegraphics[width=0.99\linewidth, trim=0.0cm 0.0cm 0.0cm 0.0cm,clip]{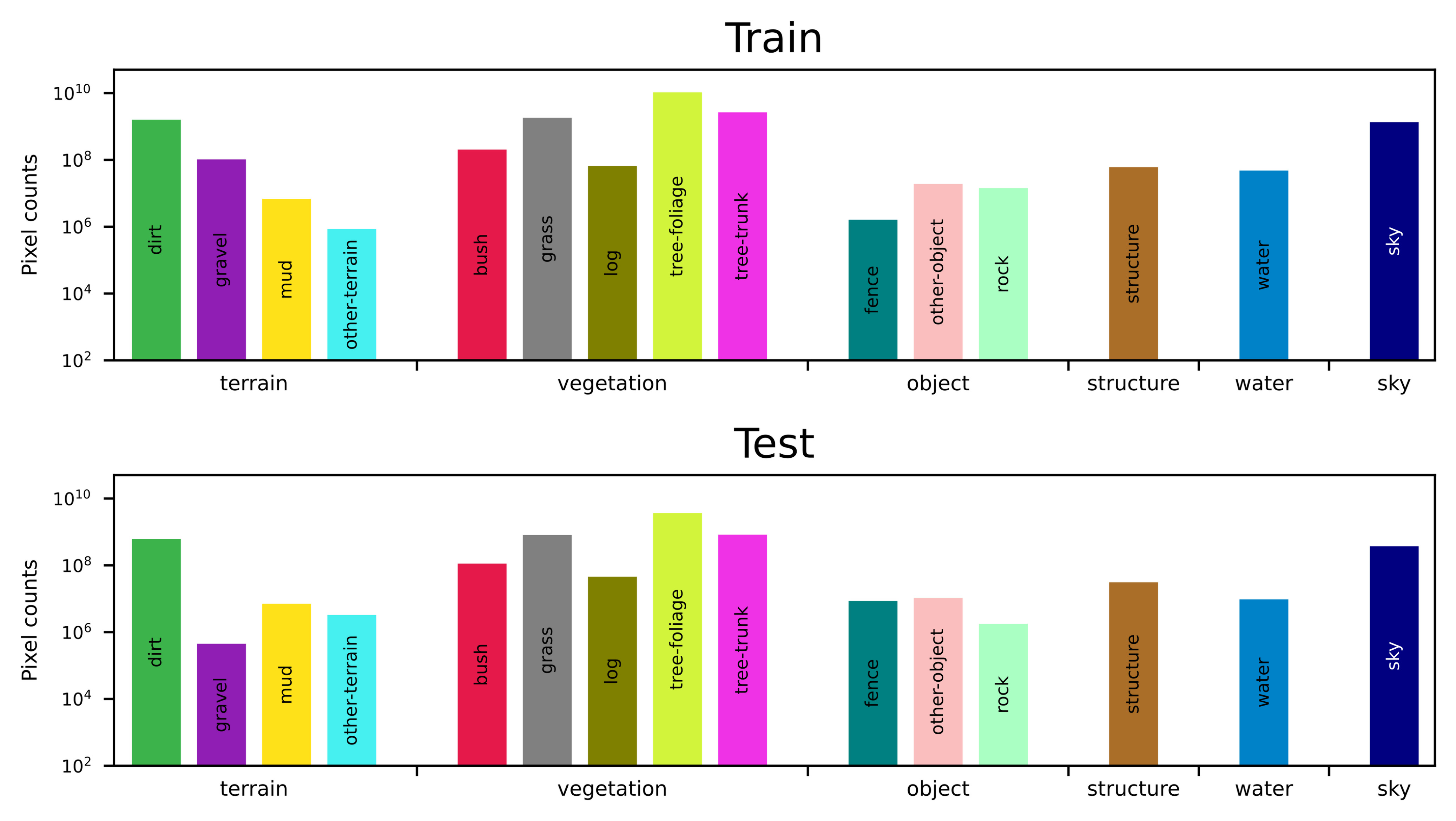}
    \end{subfigure}
    \begin{subfigure}{\columnwidth}
        \includegraphics[width=0.99\linewidth, trim=0.0cm 0.0cm 0.0cm 0.0cm,clip]{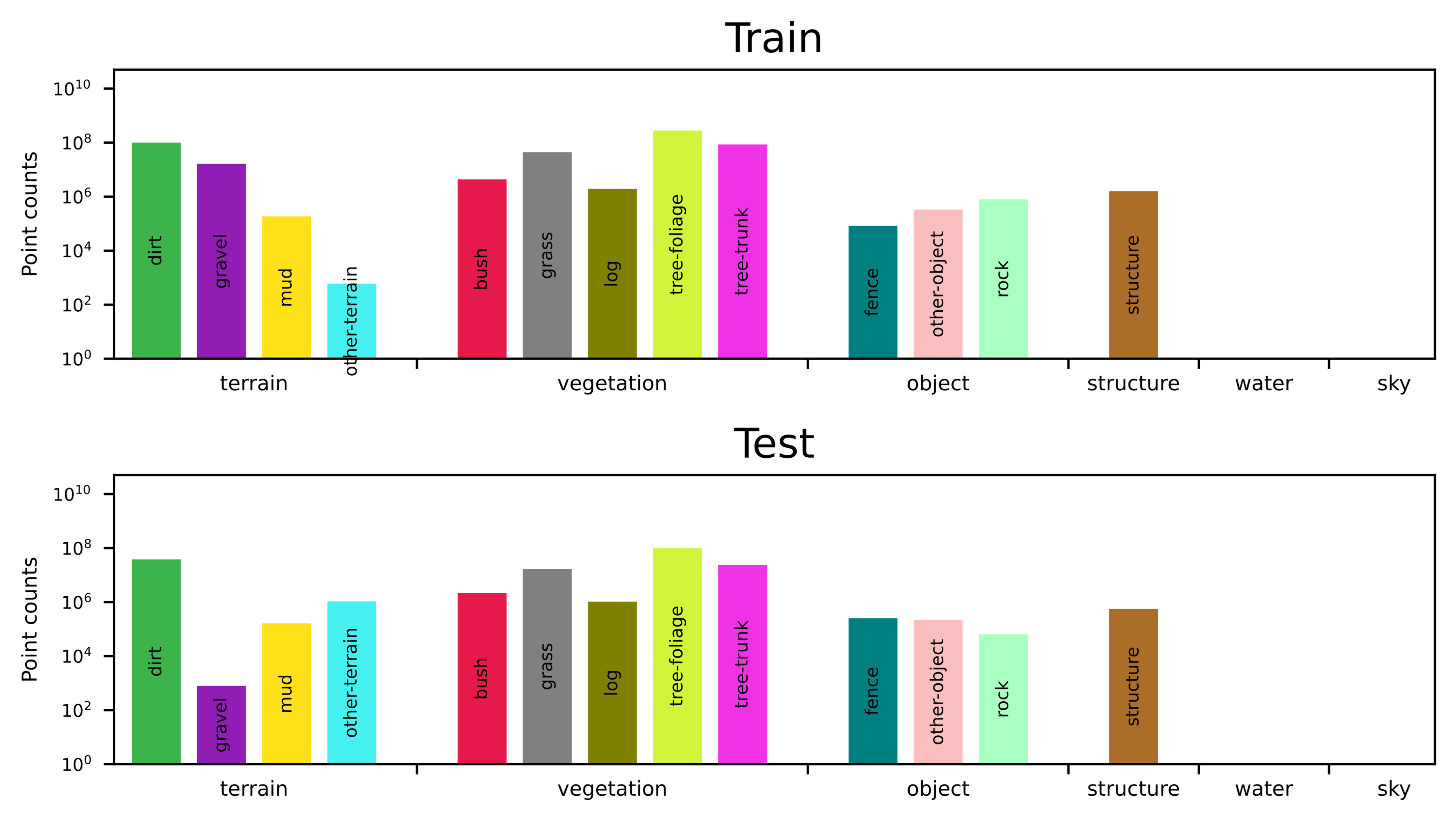}
    \end{subfigure}
    \caption{Label distributions in both 2D (top) and 3D (bottom) for our optimized train and test sets (log scale). 
    }
    \label{fig:splitlabelplot}
\end{figure}

We also show a metric map showing the distribution of our splits and their associated buffer regions between them (Figure~\ref{fig:splitplot}). Finally, in Figure~\ref{fig:splitlabelplot}, we show the label counts for each class in our optimized split.

\section{Benchmark Experiments}
\label{sec:experiments}

\subsection{Benchmark Split}

Using our split generation procedure detailed earlier, we split our dataset into train/val/test splits with $6051/283/2133$ images and $7517/356/2705$ point clouds respectively (or an objective split ratio of 70\%, 5\%, 25\%). Note that the total number of images and point clouds in these splits are less than the total contained in \coolname{}, since some images and point clouds are allocated to the buffer regions.

\subsection{2D Benchmark Experiment}
We benchmark four different approaches for 2D semantic segmentation. 
We use DeepLabv3 (\cite{chen2019rethinking}) with a Resnet-50 backbone, Mask2Former (\cite{cheng2022masked}) with a Resnet-50 backbone, Mask2Former with a Swin-L backbone, Segformer (\cite{xie2021segformer}) MiT-B5 variant, and UPerNet (\cite{xiao2018unified}) with a ConvNeXt-L backbone as our baseline methods to benchmark on our proposed dataset. These methods were chosen with different rationales. DeepLabv3 was chosen as a traditional architecture that has been commonly used in recent literature as a benchmark (\cite{strudel2021segmenter,li2022deep,cheng2021per}). As a second technique, we selected Mask2Former to provide a state-of-the-art architecture (on ADE20k~(\cite{zhou2017scene}), excluding methods with more than one billion parameters) for semantic segmentation. We selected a transformer backbone (Swin-L, pre-trained on ImageNet-22k~(\cite{deng2009imagenet})) and a convolutional backbone (Resnet-50) to investigate whether large pre-trained transformer models offer any benefits to our dataset. For our third baseline method, we selected the largest model of SegFormer (SegFormer-B5), to provide an alternate transformer segmentation network. Finally, to understand the differences between transformer and convolutional architectures on our dataset, we selected a convolutional-only method with a parameter count comparable to a large transformer architecture. Therefore, we selected Conv-NeXT-L (197 million parameters) with an UPerNet head as our fourth benchmark technique. 

\setlength\tabcolsep{3.0pt}\begin{table*}[t] %
\centering
\caption{Benchmark Semantic Segmentation on the \coolname{} Test Set. The top half of the table shows the results for 2D segmentation and the bottom half for 3D segmentation.}
\footnotesize{
\begin{tabular}{c|lc|ccccccccccccccccc}
& Method & \begin{sideways}mIoU\end{sideways} & \begin{sideways}bush\end{sideways} & \begin{sideways}dirt\end{sideways} & \begin{sideways}fence\end{sideways} & \begin{sideways}grass\end{sideways} & \begin{sideways}gravel\end{sideways} & \begin{sideways}log\end{sideways} & \begin{sideways}mud\end{sideways} & \begin{sideways}other-object\end{sideways} & \begin{sideways}other-terrain\end{sideways} & \begin{sideways}rock\end{sideways} & \begin{sideways}sky\end{sideways}  & \begin{sideways}structure\end{sideways} & \begin{sideways}tree-foliage\end{sideways} & \begin{sideways}tree-trunk\end{sideways} & \begin{sideways}water\end{sideways}\\
\midrule
\multirow{5}{*}{\hfil \textbf{2D}} & DeepLabv3 (Resnet-50) & 43.37 & 14.79 & 73.23 & 0.29 & \textbf{63.67} & \textbf{18.72} & 34.12 & 18.43 & 64.83 & 0 & 27.88 & 80.53 & 49.25 & 85.89 & 62.03 & 56.94 \\ %
& Mask2Former (Resnet-50) & 43.71 & \textbf{25.18} & 74.25 & 0.40 & 62.20 & 5.72 & 36.42 & 20.52 & 53.24 & 0 & 33.50 & 80.45 & 56.89 & 86.07 & 62.12 & 58.70 \\ %
& Mask2Former (Swin-L) & \textbf{47.85} & 23.18 & \textbf{74.67} & \textbf{0.51} & 63.01 & 7.36 & \textbf{46.85} & \textbf{21.40} & 60.59 & 0 & \textbf{60.05} & \textbf{80.57} & \textbf{64.22} & \textbf{86.15} & \textbf{63.28} & \textbf{65.97} \\ %
& Segformer (MiT-B5) & 40.83 & 13.16 & 73.32 & 0.24 & 59.97 & 7.11 & 29.64 & 2.88 & 54.67 & 0.20 & 38.30 & 79.35 & 53.01 & 84.50 & 61.42 & 54.70 \\ %
& UPerNet (ConvNeXt-L) & 47.30 & 13.78 & 75.15 & 0.35 & 63.79 & 15.61 & 39.97 & 16.85 & \textbf{69.14} & \textbf{4.96} & 55.68 & 80.48 & 60.97 & 85.98 & 62.90 & 63.95 \\ %
\midrule %
\multirow{4}{*}{\hfil \textbf{3D}} & SPVCNN  & 36.78 & 18.88 & 83.55 & \textbf{10.84} & \textbf{70.27} & 0 & 18.42 & 0.19 & 29.45 & - & 7.68 & - & 54.32 & 89.73 & 58.02 & - \\ %
& Cylinder3D  & \textbf{40.07} &  \textbf{30.61} & 82.92 & 4.81 & 69.80 & 0 & \textbf{28.12} & \textbf{6.60} & \textbf{35.08} & - & \textbf{19.39} & - & 55.16 & \textbf{89.84} & \textbf{58.54} & - \\ %
& MinkUNet  & 36.53 & 18.99 & \textbf{83.89} & 9.88 & 70.20 & 0 & 22.81 & 0.89 & 27.79 & - & 1.04 & - & \textbf{55.40} & 89.68 & 57.74 & - \\ %
& SphereFormer  & 33.97 & 12.28 & 78.00 & 7.73 & 49.65 & 0 & 27.54 & 0.69 & 28.15 & -  & 17.59 & - & 51.88 & 85.23 & 48.90 & - \\
\bottomrule
\end{tabular}}
\setlength\tabcolsep{3.0pt} %
\label{tab:benchmarking}
\end{table*}

\subsubsection{Training Procedures:}
We use the \emph{mmsegmentation} codebase\footnote{https://github.com/open-mmlab/mmsegmentation} for running all our 2D semantic segmentation benchmarks. 
For all baselines, we train
for 80k iterations, with a batch size of 40, using two Nvidia H100 GPUs. We consistently use a crop size of $(512, 512)$. We employ the augmentations: RandomResize, RandomCrop, and RandomFlip. We employ the learning rates, optimizer, and scheduling as per the defaults for each baseline method, only adjusting the scheduler to suit our batch size. We initialize the backbones of all 2D networks with pre-trained weights from either ImageNet-1k (Deeplabv3, Mask2Former Resnet, SegFormer) or ImageNet-22k (Mask2Former Swin, UPerNet ConvNeXt).

\subsection{3D Benchmark Experiment}
We benchmark four different approaches for 3D semantic segmentation. 
We utilize SPVCNN~(\cite{tang2020searching}), Cylinder3D~(\cite{zhu2021cylindrical}), MinkUNet~(\cite{choy20194d}) and SphereFormer~(\cite{lai2023spherical}).  We selected these methods due to their high performance on the common 3D semantic segmentation benchmark Semantic-KITTI (\cite{behley2019semantickitti}) and the availability of an open-source implementation.  Because LiDAR returns are non-existent or inaccurate on the `sky' and `water' classes, respectively, we exclude these classes from evaluation for our 3D benchmarking.  We also exclude the class `other-terrain' due to an inadequate number of points in the train set, leading to a total of 12 classes for the 3D benchmark. 

\subsubsection{Training Procedures:} For all 3D benchmarking, we use the \emph{mmdetection3d} codebase\footnote{https://github.com/open-mmlab/mmdetection3d} and the \emph{SphereFormer} codebase\footnote{https://github.com/dvlab-research/SphereFormer}.  
For all baselines, we train
for 50 epochs with a batch size of 20 on one or two NVIDIA H100 GPUs.
We employ the augmentations: RandomRotation, RandomScale, RandomTranslate, RandomFlip.  We employ each method's default learning rate, optimizer, and scheduling and all 3D networks are initialized with random weights.

\subsection{Evaluation Criteria}

For evaluating the performance of a semantic segmentation method with respect to the ground truth label annotations, we use the standard Mean Intersection over Union (mIoU) metric for a set of 15 classes (2D) and 12 classes (3D).

\section{Results and Discussion}
\label{sec:discussion}

\subsection{2D Semantic Segmentation}

In Table \ref{tab:benchmarking}, we display the results of our benchmark experiments on \coolname{}. Considering the task of 2D semantic segmentation, we observe that \coolname{} provides a challenging benchmark for existing techniques with a peak mIoU of $47.85$ from Mask2Former with a Swin-L backbone, closely followed by $47.3$ from UPerNet with a ConvNeXt-L backbone. Considering that M2F achieves $56.1$ (\cite{cheng2022masked}) and ConvNeXt-L achieves $53.7$ (\cite{liu2022convnet}) on ADE20k-val, it is clear that \coolname{} dataset has challenging properties for semantic segmentation.

While the mean IoU metric is important for a general performance measure, as a result of the long-tailed distribution of class frequencies in natural environments (refer to Figure.~\ref{fig:classlabeldist}), it is skewed by the presence of rarely occurring classes with a poor IoU score. Our four least common classes are \emph{other-terrain}, \emph{mud}, \emph{fence}, and \emph{rock}. Our four lowest performing classes using Mask2Former+Swin-L are \emph{other-terrain}, \emph{fence}, \emph{mud} and \emph{gravel}, with IoU scores of $0$, $0.51$, $21.4$ and $7.36$ respectively. Infrequent classes have low IoU at test time, which is a result of the lack of training samples from which the network can learn. This opens up an interesting direction for future research considering how to design or pre-train semantic segmentation networks to handle uncommon classes.

However, the number of class labels represents only part of the overall context. For example, even though \emph{gravel} only achieves an IoU of $7.36$, it is relatively commonly occurring with $10^8$ annotated pixels across the dataset. Furthermore, \emph{tree-trunk} is very common ($>10^{9}$ annotated pixels) but only achieves an IoU of $63.28$. We hypothesize that the unstructured and ambiguous properties of natural environments are additional contributing factors to the performance of semantic segmentation on specific classes.

Another observation is the low performance of the \emph{fence} class. We note that the \emph{fence} class is adequately present in both training and test sets; however, as a coincidental result of our split optimization, one design of fence (a fence with a single horizontal railing) appears in the train set while the fence in the test set has three horizontal railings. Therefore, we suspect that none of the networks can successfully generalize between different fence designs based on the existing training data.

Comparing the performance of different networks for certain classes (\eg{} \emph{dirt}, \emph{tree-foliage}, \emph{sky}, \emph{tree-trunk}, \emph{grass}), the IoU score is almost the same for any network. While for other classes (\eg{} \emph{bush}, \emph{gravel}, \emph{rock}, \emph{log}), the IoU score varies considerably between different networks. The key difference is that the aforementioned stable classes are also the top-5 most common classes in terms of pixel counts (see Figure~\ref{fig:classlabeldist}). We hypothesize that differing network architectures and pre-trained configurations have a larger influence on classes with a reduced number of training samples, resulting in the observed differences in IoU between different networks. Another observation is that Segformer is the lowest performing method (by mIoU score). One difference is that Segformer is pre-trained on ImageNet-1K (\cite{xie2021segformer}), while ConvNeXt-L and Swin-L are pre-trained on the larger ImageNet-22K, and we hypothesize that the reduced pre-training means that the network requires more training samples for the less common classes.

\setlength\tabcolsep{3.0pt}\begin{table*}[t] %
    \centering
    \caption{\coolname{} domain shift experiments. The top half of the table displays the results for the temporal domain shift (between Winter and Summer) and the bottom half represents the environmental domain shift (between Venman and Karawatha). 2D experiments are done using DeepLabv3 and 3D experiments using MinkUNet. }
    \footnotesize{
    \begin{tabular}{c|c|lc|ccccccccccccccccc}
    & Train - Test & \begin{sideways}Modality\end{sideways} & \begin{sideways}mIoU\end{sideways} & \begin{sideways}bush\end{sideways} & \begin{sideways}dirt\end{sideways} & \begin{sideways}fence\end{sideways} & \begin{sideways}grass\end{sideways} & \begin{sideways}gravel\end{sideways} & \begin{sideways}log\end{sideways} & \begin{sideways}mud\end{sideways} & \begin{sideways}other-object\end{sideways} & \begin{sideways}other-terrain\end{sideways} & \begin{sideways}rock\end{sideways} & \begin{sideways}sky\end{sideways}  & \begin{sideways}structure\end{sideways} & \begin{sideways}tree-foliage\end{sideways} & \begin{sideways}tree-trunk\end{sideways} & \begin{sideways}water\end{sideways}\\
    \midrule %
    \multirow{5}{*}{\hfil \begin{sideways}\textbf{Temporal}\end{sideways}} & Summer - Summer & \textbf{2D} & 48.51 & 17.07 & 73.63 & - & 60.40 & - & 27.86 & 0 & 61.29 & - & 22.57 & 79.56 & 53.79 & 84.91 & 58.44 & 42.58 \\ %
    & & \textbf{3D} & 30.30 & 4.76 & 80.37 & - & 47.16 & - & 15.50 & 0 & 20.94 & - & 0 & - & 0 & 84.88 & 49.35 & - \\ %
    \cmidrule{2-20}
    & Winter - Summer & \textbf{2D} & 43.95 & 16.96 & 63.88 & - & 54.54 & - & 23.16 & 0 & 39.32 & - & 41.78 & 80.03 & 56.69 & 83.30 & 59.86 & 7.87 \\  %
    & & \textbf{3D} & 27.20 & 9.68 & 82.25 & - & 45.77 & - & 0.36 & 0 & 0 & - & 0.13 & - & 0 & 84.55 & 49.28 & - \\
    \midrule %
    \midrule
    \multirow{5}{*}{\hfil \begin{sideways}\textbf{Environmental}\end{sideways}} & & & &\\
    & Karawatha - Karawatha & \textbf{2D} & 45.38 & 12.42 & 67.82 & - & 54.83 & - & 24.76 & 17.24 & 61.16 & - & 6.77 & 82.71 & 52.43 & 86.80 & 60.18 & 17.48 \\ %
    & & \textbf{3D} & 35.74 & 8.70 & 78.49 & - & 41.78 & - & 12.12 & 0 & 18.30 & - & 0.02 & - & 61.09 & 86.21 & 50.65 & - \\ %
    \cmidrule{2-20}
    & Venman - Karawatha & \textbf{2D} & 36.13 & 8.48 & 67.69 & - & 52.57 & - & 26.78 & 0.91 & 35.94 & - & 0.06 & 81.01 & 0.13 & 85.24 & 57.59 & 17.11 \\ %
    & & \textbf{3D} & 30.78 & 13.24 & 82.16 & - & 43.10 & - & 14.84 & 0 & 17.12 & - & 0 & - & 0.78 & 85.90 & 50.64 & - \\ %
    \bottomrule
    \end{tabular}}
    \setlength\tabcolsep{3.0pt} %
    \label{tab:interseq}
    \end{table*}

\subsection{3D Semantic Segmentation}

On the bottom half of Table~\ref{tab:benchmarking}, we provide the benchmark performance of 3D point cloud segmentation methods. We observe that 3D semantic segmentation is a challenging task on \coolname{}, with the highest mIoU of $40.07$ achieved by Cylinder3D (\cite{zhu2021cylindrical}). Overall, we observe that all techniques have a relatively similar mIoU; while it is expected that SPVCNN~(\cite{tang2020searching}) will perform similarly to Cylinder3D (since they have a similar mIoU on SemanticKITTI), there are little variations in mIoU between techniques. We hypothesize that the unstructured properties of natural environments limit the ability of these networks to perform semantic segmentation, with some classes being much easier to classify than others.

We observe a considerable variation in IoU between different classes in 3D. Commonly occurring classes such as \emph{dirt} and \emph{tree-foliage} achieve a consistently high IoU score across all four benchmark methods. In fact, we find that point cloud segmentation networks are more accurate at identifying dirt than 2D segmentation networks. While grass and tree trunk are also common classes (approximately $10^7$ and $10^8$ points respectively), we observe a notable drop in IoU to approximately $70\%$ and $60\%$ respectively. And yet, \emph{structure} has an IoU of approximately $55\%$ even though it is significantly less common in terms of class label counts - approximately $10^6$ 3D points. We attribute this to the distinctness of classes. As the only information received by the network is the 3D coordinates of points across 3D space, classes that have distinctive shapes and surfaces in 3D are more likely to be easy to classify (\eg{} structures, fences).

In another example, the classes \emph{bush} and \emph{log} both experience low IoU scores of around $20\%$ and yet are more commonly occurring than \emph{structure}. We hypothesize that it is likely difficult to distinguish between \emph{bush} and \emph{tree-foliage}, and between \emph{log} and \emph{tree-trunk}, as these classes are naturally ambiguous with each other, especially in 3D where there is no color information available. We suggest that an interesting avenue of future work using \coolname{} would be to use geometric projection to supplement a point cloud with color information from images, which may aid in perception in these types of unstructured natural environments.

Finally, as in 2D, we observe that tail classes that rarely occur are difficult to learn from with existing training paradigms. In 3D, our five least common classes are \emph{mud}, \emph{fence}, \emph{other-object}, \emph{other-terrain} and \emph{rock}. We find that all these classes have low IoU scores, although \emph{fence} achieves a surprisingly high IoU (relative to its counterpart in 2D) of up to $10.84\%$ with SPVCNN~(\cite{tang2020searching}), even though it is a rare class with just $10^5$ 3D points. Again, as was the case with \emph{structure}, \emph{fence} is an object type with a distinctive and structured shape in 3D, which likely provides a bias toward the ability of a network to learn to classify this class. Finally, we observe that \emph{gravel} is unable to be learned by any network, even though there are $10^7$ points in the dataset. We hypothesize that \emph{gravel}, alongside \emph{mud}, is difficult to identify in 3D due to sharing a very similar shape and environmental context to the extremely common class \emph{dirt}.%

\subsection{Impact of Temporal and Environmental Domain Shifts}

\coolname{} is comprised of both repeat traverses of the same natural environment across six months and traversals across spatially separate environments. This allows us to quantitatively evaluate the performance of our trained semantic segmentation models in the presence of a domain shift, \ie{} a shift in the data distribution between the training and testing sets as a result of either a change in location or due to changes in the natural environment over time. 

For this experiment, we generated four new train/val/test splits, which are subsets of the existing optimized split. Our splits are detailed below (the validation set always stays the same):
\begin{itemize}
    \item Summer to Summer: train and test on sequences from December (Summer season in Australia) (V-03 and K-03). Train and test sets maintain geographic separation. Train: 3742 images, Test: 1499 images.
    \item Winter to Summer: train on June (Winter season in Australia) (V-01, V-02 and K-01), test on December (V-03 and K-03). Train: 2309 images, Test: 1499 images.
    \item Karawatha to Karawatha: train on the training regions from K-01 and K-03, and test on the test regions from K-01 and K-03. Train: 3809 images, Test: 1247 images.
    \item Venman to Karawatha: train on the training regions from V-01, V-02 and V-03, and test on the test regions from K-01 and K-03. Train: 2242 images, Test: 1247 images.
\end{itemize}

\begin{figure}[t]
    \centering
    \includegraphics[width=\columnwidth]{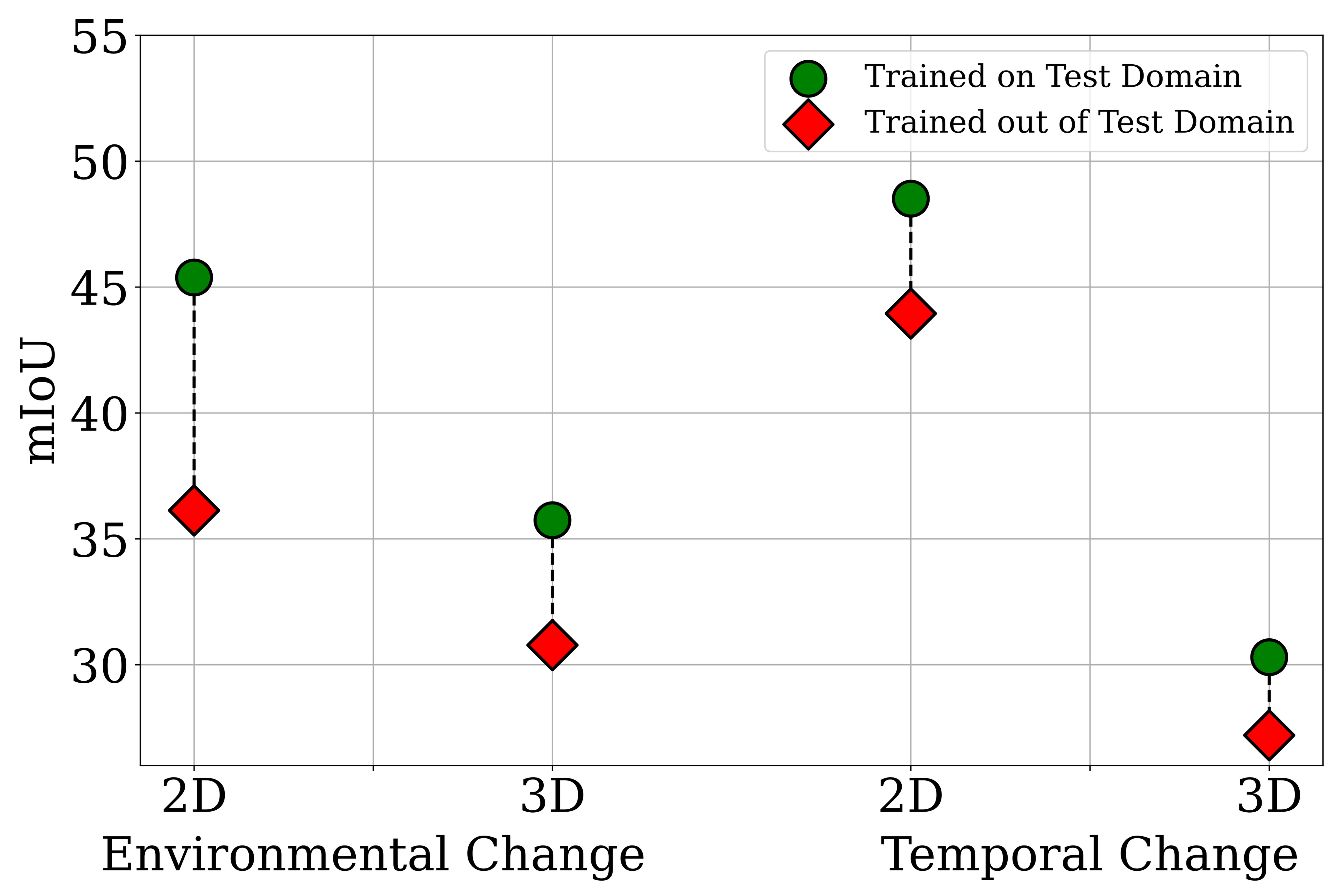}
    \caption{Visualisation of the performance drop due to environmental or temporal domain shifts between the training and testing sets. ``Environmental Change'' estimates the domain gap between Venman and Karawatha, and `Temporal Change' estimates the domain gap between Winter and Summer. ``Trained on Test Domain'' refers to experimental setups where the training and testing splits are in-domain (\eg{} Summer to Summer), whereas `Trained out of Test Domain' refers to setups where there is either an environmental or temporal domain shift between the training and testing splits (\eg{} Winter to Summer).}
    \label{fig:domain_adaption}
\end{figure}

We provide benchmarks for training and testing on these sub-splits in Table \ref{tab:interseq} for both 2D (using Deeplabv3) and 3D (using MinkUNet), and visualize the impact of the temporal and environmental domain shifts on test-time performance in Figure \ref{fig:domain_adaption}. Note that we exclude the classes \emph{gravel} and \emph{other-terrain} from evaluation on these sub-splits, due to very low class counts in either the train or test set.

\subsubsection{Temporal Domain Shift:} In 2D, we observe that semantic segmentation performs better when no temporal domain shift occurs with respect to the training data, as expected. In natural environments, it is expected that vegetation classes, especially \emph{tree-foliage}, \emph{grass}, and \emph{bush}, will change more rapidly over time than features such as \emph{rock}, \emph{structures}, \emph{objects}, and \emph{dirt}. Furthermore, vegetation can also change color due to seasonal changes. For example, it can be observed that the IoU for \emph{grass} is higher in Summer to Summer ($60.4$) than in Winter to Summer ($54.54$). In 3D, a similar but smaller trend exists. For example, the \emph{grass} class IoU increases from $45.77$ to $47.16$ when the training season is the same as the test season. This is not surprising - grass is likely to be the type of vegetation that grows the fastest and is most affected by seasonal differences. However, the inverse trend exists for the \emph{bush} class with a drop in IoU of $4.92$; although, as \emph{bush} is a rare class there may be insufficient training data (in these inter-sequence splits) for stable training of this class. 

\subsubsection{Environmental Domain Shift:} 
In addition to the above results, we observe that the mIoU drops considerably for both 2D and 3D modalities when there is an environmental domain shift between the training and testing data. Some classes are highly impacted by the environment used for training. For example, structures are only able to be detected in the Karawatha test set when Karawatha is also used for training - this indicates that the types of structures in Venman are very different in their style/design and are unable to generalize to structures in Karawatha (in both 2D and 3D).

A number of classes also appear to be invariant to the physical location of the training set. We observe that \emph{dirt}, \emph{grass}, \emph{log}, \emph{sky}, \emph{tree-foliage}, \emph{tree-trunk} and \emph{water} are almost unchanged in their IoU when the training environment changes from Karawatha to Venman. This is an expected result since these classes comprise features that are commonly found in natural environments. However, we note that since both environments are located in Australia, we would expect a greater impact on IoU if a training or testing set from a natural environment in another country was used. This would be an interesting avenue for future work.

\subsection{Label Histograms}

As discussed earlier, our 3D annotation procedure allows us to provide label distributions for every 3D point. We provide a histogram of the number of times a given class was assigned to that point from all 2D semantic labels (observations from human annotators across multiple frames) of that point. We propose that natural environments are naturally ambiguous and do not easily conform to rigid semantic label assignments. For example, the difference between \emph{dirt} and \emph{mud} is small since mud is simply wet dirt. Or, a small tree could be confused with a bush/shrub. In this section, we analyze which classes are co-occurring in the label histograms to understand which pairs of classes are naturally ambiguous with each other.

\begin{figure}[t]
    \centering
    \begin{subfigure}[b]{0.46\columnwidth}
        \includegraphics[width=\linewidth, trim=5.0cm 0.0cm 1.0cm 0.0cm,clip]{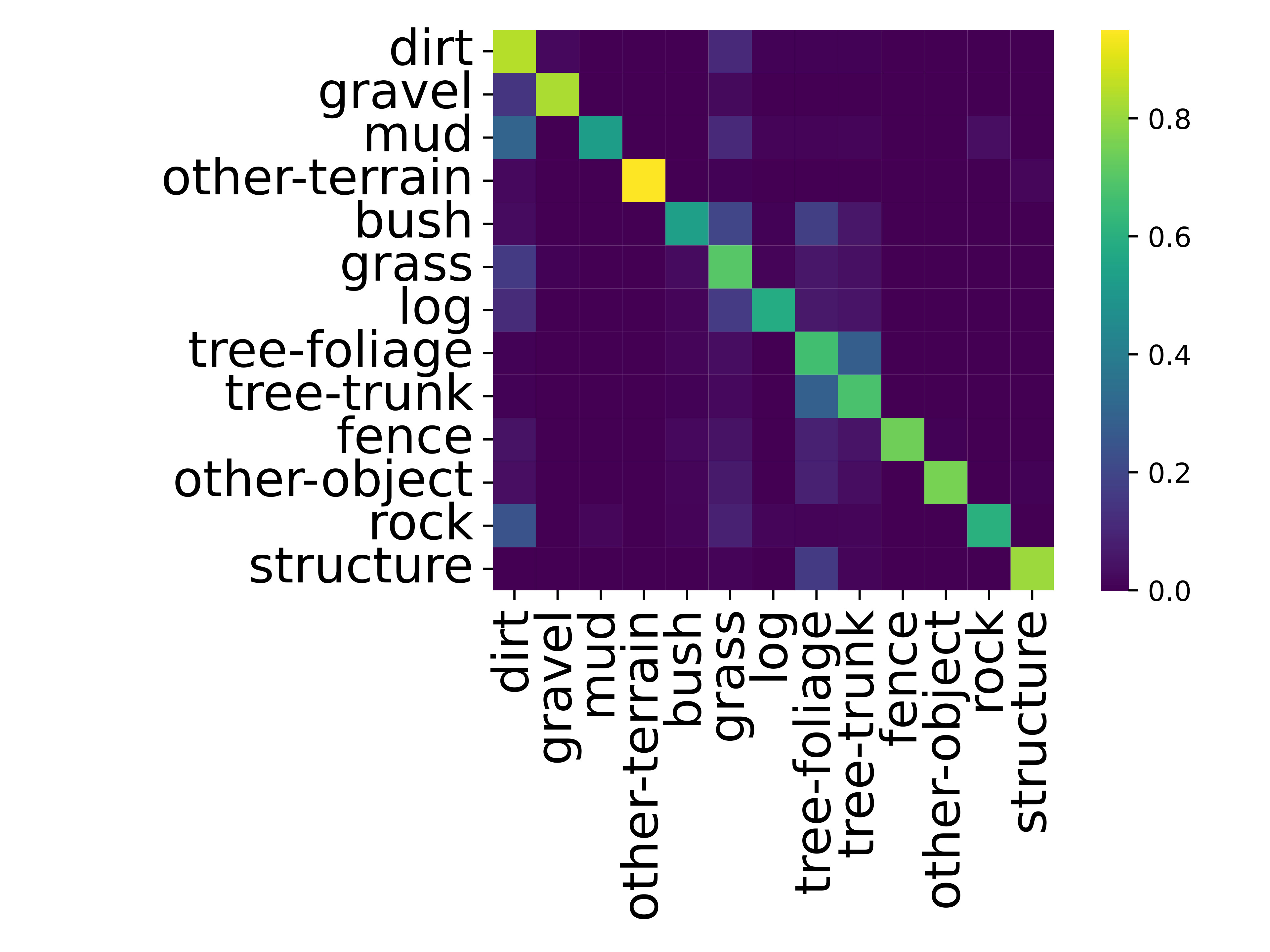}
    \end{subfigure}
    \hfill
    \begin{subfigure}[b]{0.4\columnwidth}
        \includegraphics[width=\linewidth, trim=20.0cm 80.0cm 20.0cm 5.0cm,clip]{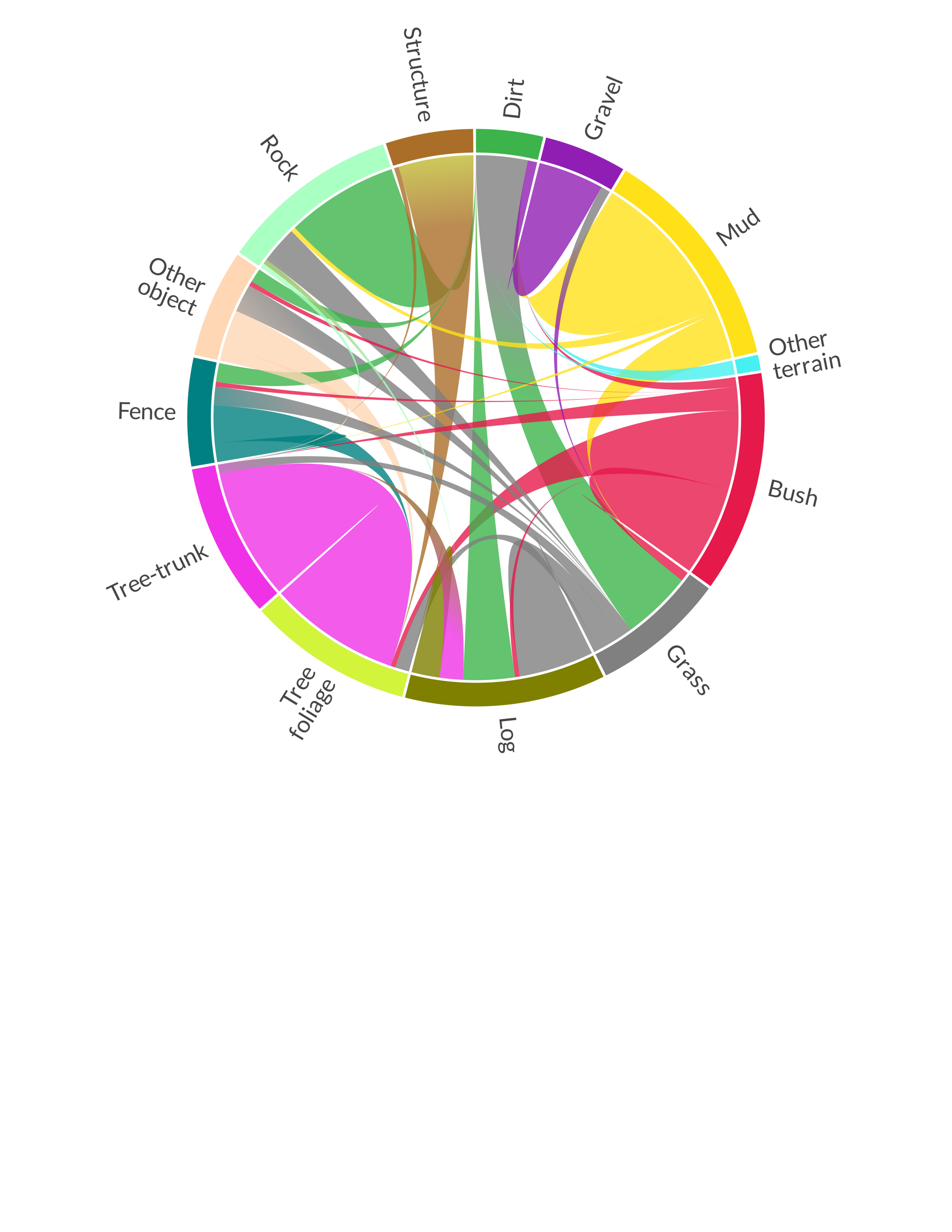}
    \end{subfigure}
    \hfill
    \caption{Utilizing the histograms of labels produced by label transfer, we can calculate the co-occurrences of different classes, \ie{} for a given 3D point, what labels were assigned from all of its viewpoints. Left: we plot the co-occurrences in matrix form, with all rows normalized to sum to one. Right: we represent co-occurrences via a chord diagram. A larger outer segment denotes a class with greater co-occurrences with other classes, and the directions of the curves denote which pairs of classes are co-occurring.}
    \label{fig:amb_plots}
\end{figure}

To measure the semantic ambiguity - the inconsistency in annotation due to the natural ambiguity of natural environments - we calculate co-occurrences between classes in the histograms. We calculate the co-occurrences for each 3D point, then aggregate across all 3D points in our dataset. Our results are shown in Figure~\ref{fig:amb_plots}, representing the same data in both matrix form and using a chord diagram. In the matrix, a larger value of a class diagonal denotes a less ambiguous class, \ie{} a class where all 2D observations had the same label. In the chord diagram, the arrows signify which pairs of classes are co-occurring with each other.

From these plots, we can make the following conclusions. First, a large proportion of \emph{mud} points have also been labeled as \emph{dirt}, however conversely, only a small fraction of dirt points have been given mud labels in 2D (noting that there are approximately $10^5$ \emph{mud} points versus $10^8$ \emph{dirt} points). \emph{Tree-trunk} and \emph{tree-foliage} are also co-occurring, however, this is not unexpected - \coolname{} contains sections of dense forest trails where a fine-grained segmentation between tree leaves and branches becomes ill-defined, especially from a distance.  However, overall, the mean value of the diagonal of the co-occurrence matrix is $0.71$. Therefore, the total disparity in 2D observations is still small, which is a result of our fine-grained annotation auditing procedure. We provide these label histograms with our benchmark dataset release, which could be used in future work such as uncertainty-aware semantic segmentation in natural environments.

\section{Conclusion}
\label{sec:conclusion}

In this paper, we have introduced \coolname{}, a new benchmark dataset for 2D and 3D semantic segmentation in natural environments. \coolname{} comprises traverses across multiple different natural forest environments over an extended time period and provides high-resolution 2D images and dense 3D LiDAR point clouds with full point / pixel level annotation. Additionally, we use LiDAR SLAM to provide accurate 6-DoF pose information for all camera and LiDAR submap timestamps. \coolname{} comprises 9,306 annotated images and 12,148 annotated 3D point cloud submaps, across 21km of walking through densely vegetated natural environments. The annotation divides the natural environment into 15 classes, classifying both different vegetation types (\eg{} bushes/shrubs versus trees) and different terrain types (\eg{} dirt versus gravel), along with other features including fences and structures.

We provide an initial benchmark using state-of-the-art 2D and 3D segmentation methods on \coolname{}, to demonstrate the additional challenges present in unstructured natural environments. We demonstrate that \coolname{} poses challenges for existing segmentation methods in both 2D and 3D, as a result of the inherent semantic ambiguity and long-tail distribution of class occurrences in natural environments.  In addition, by providing accurate 6-DoF pose information for both the image and LiDAR modalities we open up the opportunity for future researchers to investigate multi-modal segmentation approaches which can leverage the advantages of both modalities - the rich color and textural information from RGB images as well as the 3D geometric structure provided by the LiDAR point cloud.

We expect that \coolname{} will aid in the development of future perception systems for autonomy for applications such as search and rescue, conservation, and agricultural automation, for which existing urban perception datasets are ill-suited. Future work includes developing novel methods for semantic segmentation in both 2D and 3D, designed specifically for segmentation in natural environments. Additional future work also includes expanding this dataset to include instance annotations. For example, the inclusion of instance annotations for different trees would provide value in training networks for tree detection and classification. Combining all of these aspects, we believe \coolname{} provides a valuable resource for the future development of semantic segmentation techniques for robust autonomous perception in natural environments.

\begin{acks}
The authors gratefully acknowledge funding of the project by the CSIRO’s Machine Learning and Artificial Intelligence (MLAI) FSP and continued support from the CSIRO's Data61 Embodied AI Cluster. This work would not be possible without support from members of the CSIRO Robotics including Brett Wood, Dennis Frousheger, Nick Hudson, Paulo Borges, Gavin Catt, Fred Pauling, Dave Haddon, and Stano Funiak.
\end{acks}

\newpage

\section{\coolname{} Supplementary Material}
\subsection{Detailed Class List}

To ensure effective training of neural networks on \coolname{}, we performed post-processing on our class list (as annotated) to ensure that the set of classes had a similar label count. We defined an exclusion threshold of $10^6$ pixels; that is, any class that has less than $10^6$ pixels would be excluded from evaluation.

To maximize the utility of \coolname{}, it is released with raw annotation files containing a superset of 18 classes. Of these 18 classes, we merged \emph{pole} into \emph{other-object}, merged \emph{asphalt} into \emph{other-terrain}, and excluded \emph{vehicle} from evaluation - leaving an evaluation set of 15 classes for 2D semantic segmentation. These decisions were based on our exclusion threshold and the logical semantic grouping of \emph{asphalt} into \emph{other-terrain} and \emph{pole} into \emph{other-object}. Additionally, because LiDAR returns are either non-existent or inaccurate for \emph{sky} and \emph{water}, respectively, our evaluation set for 3D segmentation does not include \emph{sky} and \emph{water}. We also remove \emph{other-terrain} due to low point counts in our splits; therefore, our 3D evaluation set contains 12 classes. Our dataset code repository contains the necessary functions to automatically merge these additional classes at run-time, however the option exists to utilize these classes. In Table~\ref{tab:semdesc}, we provide a full description of all 18 classes and their semantic ontology.

\begin{figure}[h!]
    \centering
    \includegraphics[width=\columnwidth, trim=0.0cm 0.4cm 0.4cm 0.0cm,clip]{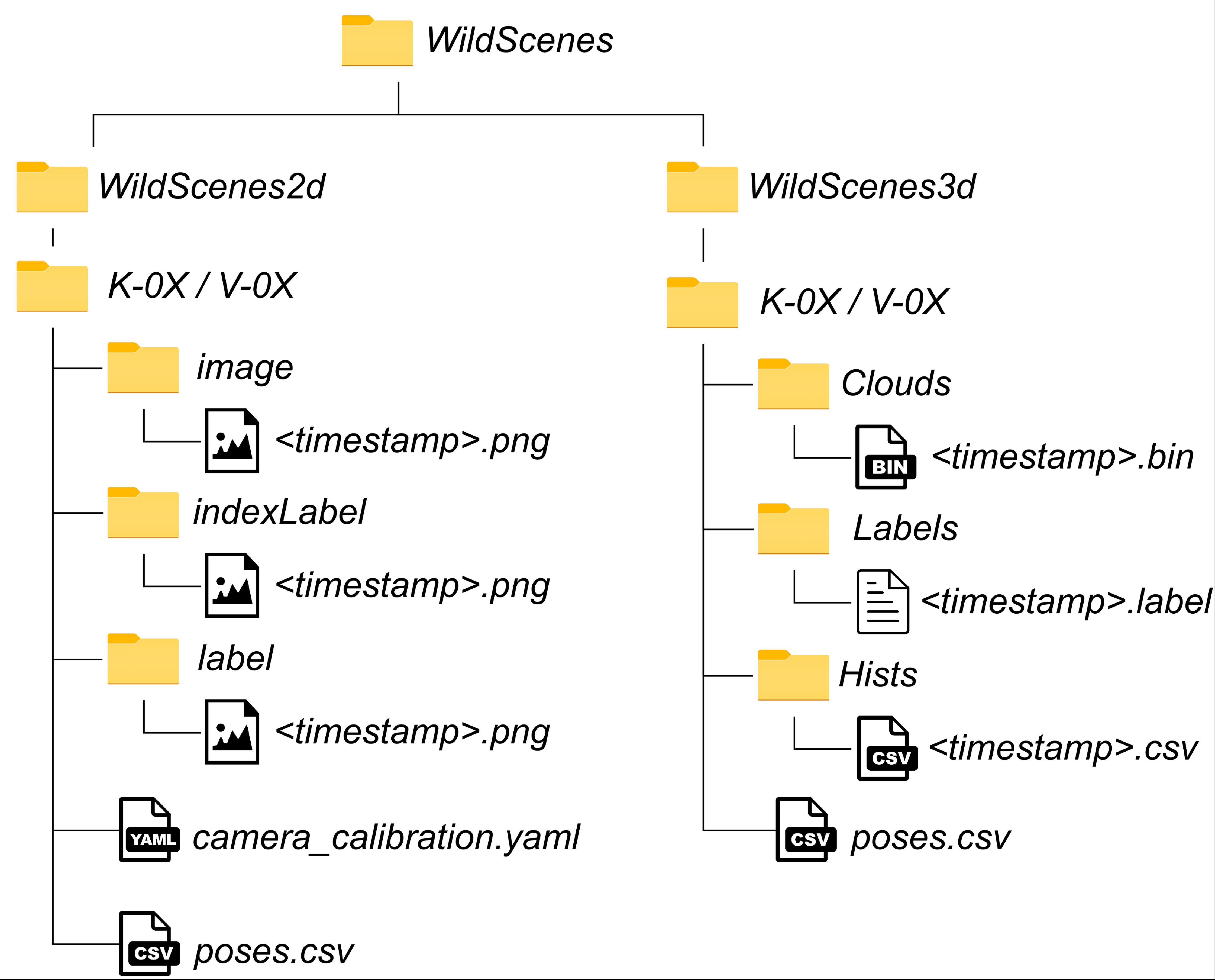}
    \caption{File structure of the \coolname{}, illustrating the organizations of images, LiDAR scans, labels, and pose / camera calibration files for each sequence.}
    \label{fig:file_structure}
\end{figure}

\subsection{File Formats}
Figure \ref{fig:file_structure} provides an outline of the file structure of the \coolname{} benchmark dataset.  

In Table~\ref{tab:fileformats}, we provide a description of each subfolder and file type in our dataset. All 2D image and 3D point cloud files, along with their corresponding labels are named with a timestamp that can be used to identify 2D-3D correspondences.

\subsection{Qualitative Results}

\begin{figure*}[t]
    \centering
    \includegraphics[width=0.99\linewidth, trim=0.0cm 80cm 0.0cm 0.0cm,clip]{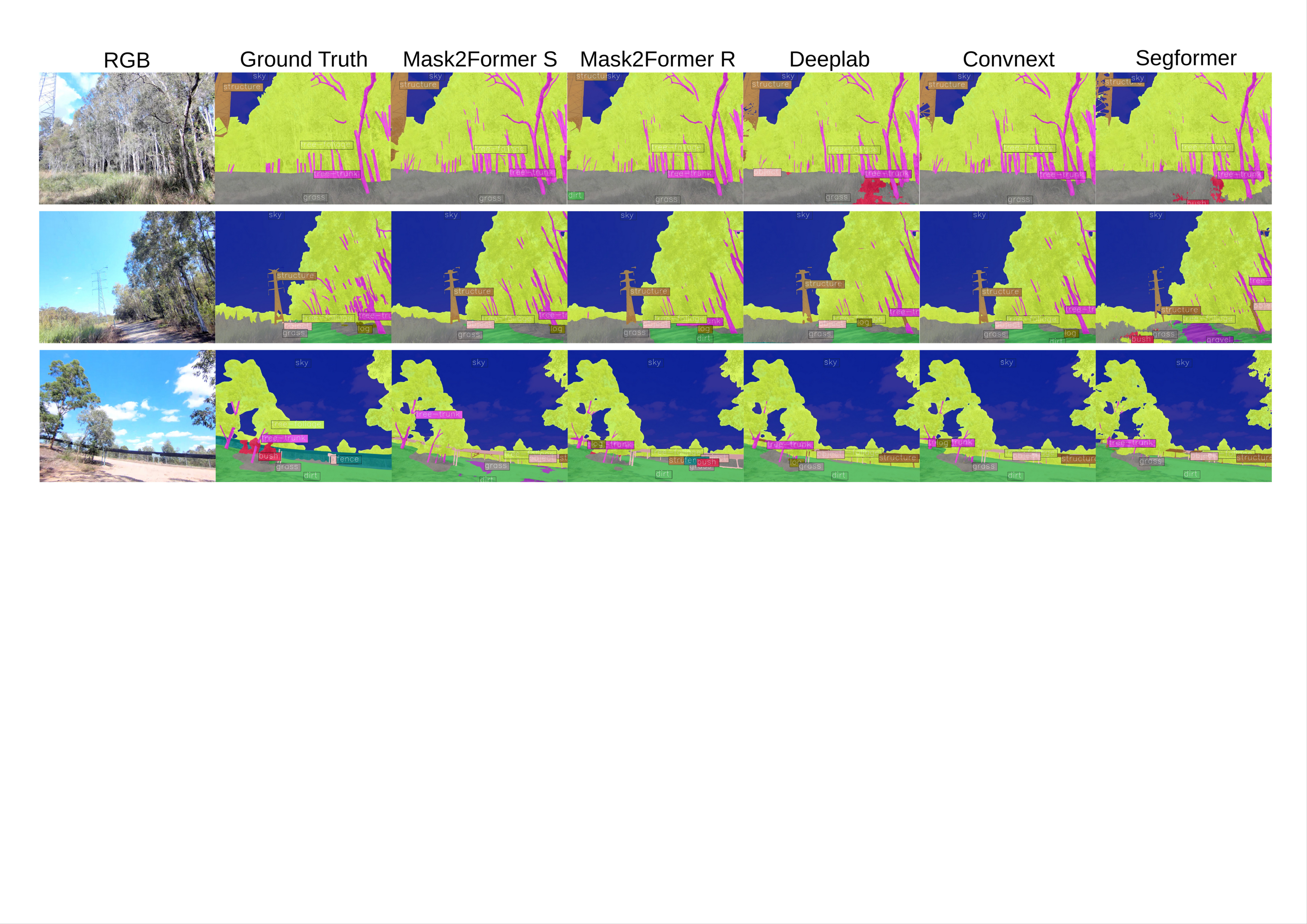} %
    \caption{Qualitative results on \coolname{} 2D data, comparing the semantic segmentation performance from the different methods with the ground truth labels.}
    \label{fig:qual2d}
\end{figure*}

In Figure~\ref{fig:qual2d}, we provide qualitative results of the different 2D benchmark semantic segmentation methods on \coolname{}. We observe reasonably good performance across all methods, even though the segmentation of natural environments can be very challenging. For example, the segmentation of thin branches from tree leaves and sky is challenging and in some instances, the network detects fine annotation details missed by the human annotators. We also observe some common segmentation failure cases shared amongst all methods. For instance, in the third row all methods incorrectly annotate the fence (the dark green color in the ground truth). The existing methods either classify the fence as an object, or as a structure. Given that a fence can be considered similar to a structure (both human made), we find that the methods struggle to successfully differentiate between these classes.

\begin{figure*}[b]
    \centering
    \includegraphics[width=\linewidth, trim=1.0cm 0cm 1.0cm 0.0cm,clip]{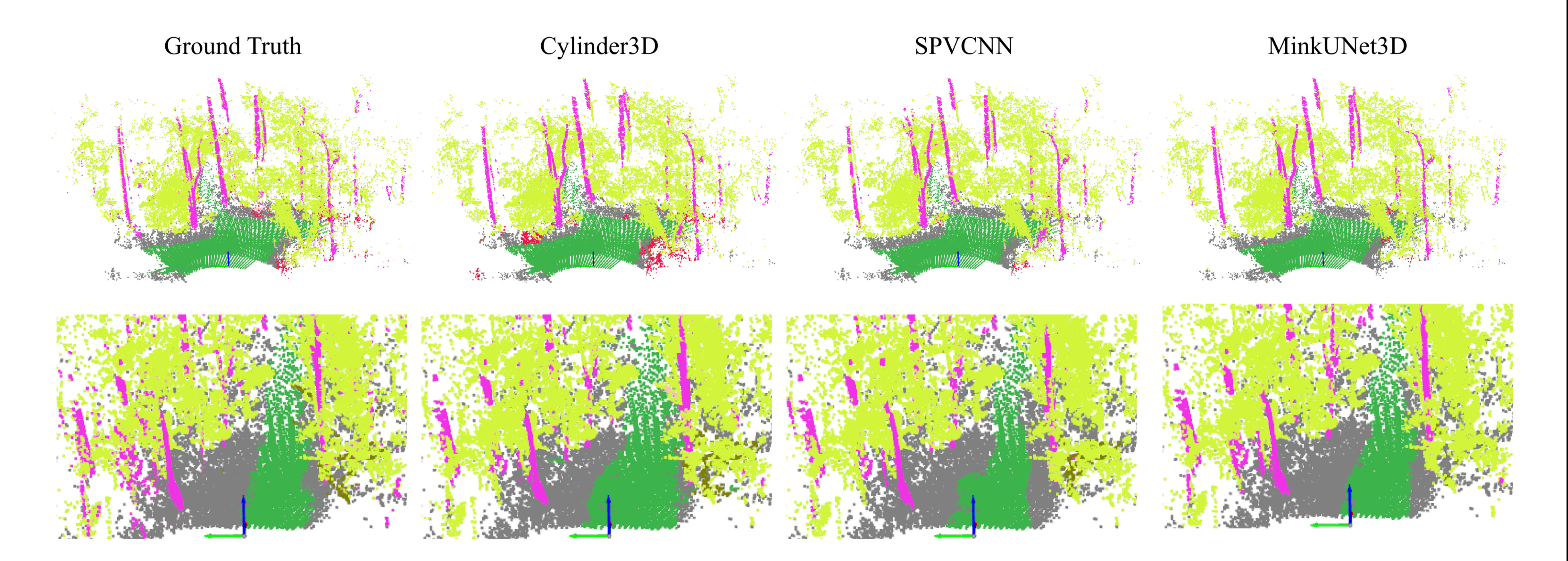}
    \caption{Qualitative results on \coolname{} 3D data, comparing the semantic segmentation performance of Cylinder3D, SPVCNN and MinkUNet with the ground truth labels.}
    \label{fig:qual3d}
\end{figure*}

In Figure~\ref{fig:qual3d}, we show qualitative results for the methods Cylinder3D, SPVCNN and MinkUNet on \coolname{}. Even on these challenging natural point clouds, we observe that very common classes such as tree-foliage are annotated very well with respect to the ground truth annotations. Some errors can be oberseved in these two examples. For example, Cylinder3D incorrectly labels some of the tree-foliage (yellow color) as bushes (red) in the example in the top row. Second, all the methods have difficulty identifying a small sign (an object, color coded light pink in the ground truth in the top row example) in the scene. Nonetheless, overall the 3D segmentation methods we benchmarked are able to correctly annotate the point clouds with a high degree of similarity to the ground truth.

\subsection{Ablation Study: Effect of Pre-training}

In our benchmark experiments, all 2D methods are initialized with pre-trained weights from ImageNet before fine-tuning on \coolname{}. In 3D, all semantic segmentation methods are initialized with random weights. In this study we experiment with different network initialization - specifically, training from random weights in 2D and trained from pre-trained weights in 3D. For this study we used the Deeplabv3 and SPVCNN networks in 2D and 3D respectively.

\begin{table}[ht!]
    \caption{Description of each file type in our \coolname{} benchmark dataset. 
    }
    \begin{center}
    \begin{tabular}{>{\centering\arraybackslash}p{0.40\linewidth}  p{0.5\linewidth}}
     \textbf{File/Folder Name} & \textbf{Description} \\
     \hline
    image & Sampled and rectified .png RBG images ($2016 \times 1512$ resolution)\\
    indexLabel & Raw annotated images in .png format. Pixel values are label index values 
    (class indices [0-14] as assigned with classes sorted alphabetically by class name). 
    \\
    label (2D) & Palette annotated images in .png format (human readable label images) \\
    Clouds & Point cloud submaps stored in .bin format and the same format as SemanticKITTI (\cite{behley2019semantickitti}).
    Point cloud are in the frame of reference of the LiDAR. 
    \\
    Labels (3D) & 3D labels are stored as .label files, again in the same format as SemanticKITTI \\
    Hists & .csv files containing the full histogram of 2D label observations for every 3D point \\
    poses.csv & .csv file containing timestamps and associated 6-DoF poses as calculated by SLAM \\
    camera\_calibration.yaml & Sensor platform intrinsics and extrinsics, one .yaml file for each sequence of the dataset \\
    \hline
    \end{tabular}
    \label{tab:fileformats}
    \end{center}
    \vspace{-0.75cm}
    \end{table}

In Table~\ref{tab:pretrain_abl}, we show the results from these ablations, comparing Deeplabv3 trained with ImageNet weights against random weights, and SPVCNN trained with SemanticKITTI (\cite{behley2021towards}) weights versus random weights. While SemanticKITTI is a different dataset with a different class list, we initialize the backbone of SPVCNN with weights trained on SemanticKITTI and use a decoder head which is specific to \coolname{}. In 2D, we observe that
removing the pre-trained ImageNet weights reduces the performance of the network after training, as expected.
In 3D, we observe the same phenomenon, with SemanticKITTI pre-training having a postive impact
on the performance of the network after training. While SemanticKITTI is an urban dataset with a different class list, we find that initializing the weights before training on \coolname{} is slightly more effective than using randomly initialized weights.

\setlength\tabcolsep{3.0pt}\begin{table*}[h] %
\centering
\caption{Ablation over different pre-training configurations on the \coolname{} Test Set. We investigate the difference between a pre-trained network and a randomly initialized network in both 2D and 3D semantic segmentation.}
\resizebox{\textwidth}{!}{
\begin{tabular}{c|lc|ccccccccccccccccc}
& Method & \begin{sideways}mIoU\end{sideways} & \begin{sideways}bush\end{sideways} & \begin{sideways}dirt\end{sideways} & \begin{sideways}fence\end{sideways} & \begin{sideways}grass\end{sideways} & \begin{sideways}gravel\end{sideways} & \begin{sideways}log\end{sideways} & \begin{sideways}mud\end{sideways} & \begin{sideways}other-object\end{sideways} & \begin{sideways}other-terrain\end{sideways} & \begin{sideways}rock\end{sideways} & \begin{sideways}sky\end{sideways}  & \begin{sideways}structure\end{sideways} & \begin{sideways}tree-foliage\end{sideways} & \begin{sideways}tree-trunk\end{sideways} & \begin{sideways}water\end{sideways}\\
\midrule
\multirow{2}{*}{\hfil \textbf{2D}} & DeepLabv3 (random weights) & 38.37 & 24.28 & 71.97 & 0.18 & 61.74 & 4.05 & 26.78 & 6.46 & 59.03 & 0 & 11.45 & 80.20 & 45.77 & 85.81 & 61.77 & 36.00 \\ %
& DeepLabv3 (ImageNet weights) & 43.37 & 14.79 & 73.23 & 0.29 & 63.67 & 18.72 & 34.12 & 18.43 & 64.83 & 0 & 27.88 & 80.53 & 49.25 & 85.89 & 62.03 & 56.94 \\ %
\midrule %
\multirow{2}{*}{\hfil \textbf{3D}} & SPVCNN (random weights) & 36.78 & 18.88 & 83.55 & 10.84 & 70.27 & 0 & 18.42 & 0.19 & 29.45 & - & 7.68 & - & 54.32 & 89.73 & 58.02 & - \\ %
& SPVCNN (SemKITTI weights) & 37.61 & 17.79 & 83.95 & 10.03 & 70.50 & 0 & 19.69 & 0.96 & 38.48 & - & 6.62 & - & 55.58 & 89.68 & 57.99 & - \\ %
\bottomrule
\end{tabular}}
\setlength\tabcolsep{3.0pt} %
\label{tab:pretrain_abl}
\end{table*}

\newpage

\begingroup
    \renewcommand{\arraystretch}{2.0} %
    
    \begin{table*}[b]
    \caption{Semantic definitions for the superset of 18 classes that were annotated. The labels denoted with a tick comprise the 15 classes evaluated in our benchmark experiments. Our exclusion threshold of $10^6$ results in the decision to exclude the \emph{vehicle} label and merge \emph{asphalt} into \emph{other-terrain}, while the similar semantic definition of \emph{pole} and \emph{other-object} in addition to the low label count motivates our decision to merge \emph{pole} into \emph{other-object}.}
    \begin{center}
    \begin{tabular}{p{0.15\linewidth} | >{\centering\arraybackslash}p{0.11\linewidth} | >{\centering\arraybackslash}p{0.1\linewidth} | p{0.55\linewidth}}%
     \textbf{Label name} & \textbf{Num. Pixels} & \textbf{Incl./Excl.} & \textbf{Definition} \\
     \hline
    \fcolorbox{black}{dirt}{\rule{0pt}{5pt}\rule{5pt}{0pt}}  Dirt  & $2.5*10^9$ & \cmark & Hard packed dirt terrain, including dirt covered by leaves. \\
    \fcolorbox{black}{gravel}{\rule{0pt}{5pt}\rule{5pt}{0pt}} Gravel & $1.1*10^8$ & \cmark & Terrain composed of loose gravel, such as gravel paths. \\
    \fcolorbox{black}{mud}{\rule{0pt}{5pt}\rule{5pt}{0pt}} Mud & $1.5*10^7$ & \cmark & Any muddy terrain, which is often found in creek gullies or adjacent to bodies of water. \\
    \fcolorbox{black}{other-terrain}{\rule{0pt}{5pt}\rule{5pt}{0pt}} Other-terrain & $4.4*10^6$ & \cmark & Any terrain that does not fall into the other categories, including terrain such as wooden bridges. \\
    \fcolorbox{black}{black}{\rule{0pt}{5pt}\rule{5pt}{0pt}} Asphalt & $2.4*10^5$ & \xmark & Asphalt or concrete terrain, and which is always human-made. Merged into Other-terrain. \\
    \fcolorbox{black}{bush}{\rule{0pt}{5pt}\rule{5pt}{0pt}} Bush & $3.4*10^8$ & \cmark & Any bush, shrub or fern plant that is larger than grass, but does not have the structure of a tree. \\
    \fcolorbox{black}{grass}{\rule{0pt}{5pt}\rule{5pt}{0pt}} Grass & $3.0*10^9$ & \cmark & Any grass areas, including both cut grass and tall grasses. \\
    \fcolorbox{black}{log}{\rule{0pt}{5pt}\rule{5pt}{0pt}} Log & $1.2*10^8$ & \cmark & Large branches that are no longer standing and are lying on the ground. \\
    \fcolorbox{black}{tree-foliage}{\rule{0pt}{5pt}\rule{5pt}{0pt}} Tree-foliage & $1.6*10^{10}$ & \cmark & All parts of a tree that is not the trunk, including leaves, indistinct branches, and any far-away trees that blur into the background vegetation. \\
    \fcolorbox{black}{tree-trunk}{\rule{0pt}{5pt}\rule{5pt}{0pt} } Tree-trunk & $3.9*10^9$ & \cmark & Both the main body of all tree trunks, and all clearly observable branches. \\
    \fcolorbox{black}{fence}{\rule{0pt}{5pt}\rule{5pt}{0pt}} Fence & $1.0*10^7$ & \cmark & Fences of any style, including wooden, chainlink and steel fences. \\
    \fcolorbox{black}{object}{\rule{0pt}{5pt}\rule{5pt}{0pt}} Other-object & $2.8*10^7$ & \cmark & A combined class for any miscellaneous objects, including signs, termite nests and chairs. \\
    \fcolorbox{black}{black}{\rule{0pt}{5pt}\rule{5pt}{0pt}} Pole & $3.5*10^6$ & \xmark & Vertical pole objects of any size, including signpoles and roadsigns. Merged into Other-object. \\
    \fcolorbox{black}{rock}{\rule{0pt}{5pt}\rule{5pt}{0pt}} Rock & $1.6*10^7$ & \cmark & Any rocks larger than pebbles, either human-made or natural. \\
    \fcolorbox{black}{black}{\rule{0pt}{5pt}\rule{5pt}{0pt}} Vehicle & $2.3*10^5$ & \xmark & Vehicles of any variety, including cars, trucks and motorbikes. Excluded from evaluation. \\
    \fcolorbox{black}{structure}{\rule{0pt}{5pt}\rule{5pt}{0pt}} Structure & $9.8*10^7$ & \cmark & Any human-made structure, including shelter such as houses, sheds and gazebos. Also includes powerline towers. \\
    \fcolorbox{black}{water}{\rule{0pt}{5pt}\rule{5pt}{0pt}} Water & $5.8*10^7$ & \cmark & Any bodies of water, including creeks, rivers and lakes. \\
    \fcolorbox{black}{sky}{\rule{0pt}{5pt}\rule{5pt}{0pt}} Sky & $2.0*10^9$ & \cmark & Any sky, also including any sections of sky in between tree foliage. Very small regions of sky in between tree leafs are classed as Tree-foliage. \\
    \hline
    \end{tabular}
    \label{tab:semdesc}
    \end{center}
    \vspace{-0.75cm}
    \end{table*}
    
    \endgroup

\clearpage
\balance{}
\bibliographystyle{SageH}
\bibliography{main.bbl}
\end{document}